\documentclass[10pt,twocolumn,letterpaper]{article}

\usepackage[pagenumbers]{cvpr}      % To produce the REVIEW version
\usepackage{graphicx}
\usepackage[numbers]{natbib}
\usepackage[bb=boondox]{mathalfa}
\usepackage{amsmath}
\usepackage{amssymb}
\usepackage{booktabs}
\usepackage{multirow}
\usepackage{makecell}
\usepackage{subcaption}
\usepackage[dvipsnames, table]{xcolor}

\usepackage[utf8]{inputenc}
\usepackage{algorithm}
\usepackage{algpseudocode}
\definecolor{cvprblue}{rgb}{0.21,0.49,0.74}

\usepackage[pagebackref,breaklinks,colorlinks,citecolor=cvprblue]{hyperref}

\def\paperID{78} % *** Enter the Paper ID here
\def\confName{CVPR}
\def\confYear{2024}

\title{ParZC: Parametric Zero-Cost Proxies for Efficient NAS}

\author{Peijie Dong$^{1 \dag}$\quad Lujun Li$^{2 \dag}$\quad Xinglin Pan$^{1}$\quad Zimian Wei$^{3}$ \quad Xiang Liu$^{1}$ \quad Qiang Wang$^{4}$\quad Xiaowen Chu$^{5}\thanks{Corresponding author, $\dag$ equal contribution.}$\\
$^{1,5}$ HKUST(GZ), $^2$ HKUST,
$^3$ NUDT, 
$^4$ HITSZ\\
{\tt\small $^{1}$\{pdong212, xpan413, xliu886\}@connect.hkust-gz.edu.cn, $^{2}$lilujunai@gmail.com,} \\
{\tt\small $^{3}$weizimian16@nudt.edu.cn,
$^{4}$qiang.wang@hit.edu.cn,
$^{5}$xwchu@ust.hk},
}

\begin{document}
\maketitle
\begin{abstract}
Recent advancements in Zero-shot Neural Architecture Search (NAS) highlight the efficacy of zero-cost proxies in various NAS benchmarks. Several studies propose the automated design of zero-cost proxies to achieve SOTA performance but require tedious searching progress. Furthermore, we identify a critical issue with current zero-cost proxies: they aggregate node-wise zero-cost statistics without considering the fact that not all nodes in a neural network equally impact performance estimation. Our observations reveal that node-wise zero-cost statistics significantly vary in their contributions to performance, with each node exhibiting a degree of uncertainty. Based on this insight, we introduce a novel method called Parametric Zero-Cost Proxies (ParZC) framework to enhance the adaptability of zero-cost proxies through parameterization. To address the node indiscrimination, we propose a Mixer Architecture with Bayesian Network (MABN) to explore the node-wise zero-cost statistics and estimate node-specific uncertainty. Moreover, we propose DiffKendall as a loss function to directly optimize Kendall's Tau coefficient in a differentiable manner so that our ParZC can better handle the discrepancies in ranking architectures. Comprehensive experiments on NAS-Bench-101, 201, and NDS demonstrate the superiority of our proposed ParZC compared to existing zero-shot NAS methods. Additionally, we demonstrate the versatility and adaptability of ParZC by transferring it to the Vision Transformer search space. 
\end{abstract}

\section{Introduction}
\label{sec:intro}

\begin{figure}[t]
    \centering
    \includegraphics[width=1.0\linewidth]{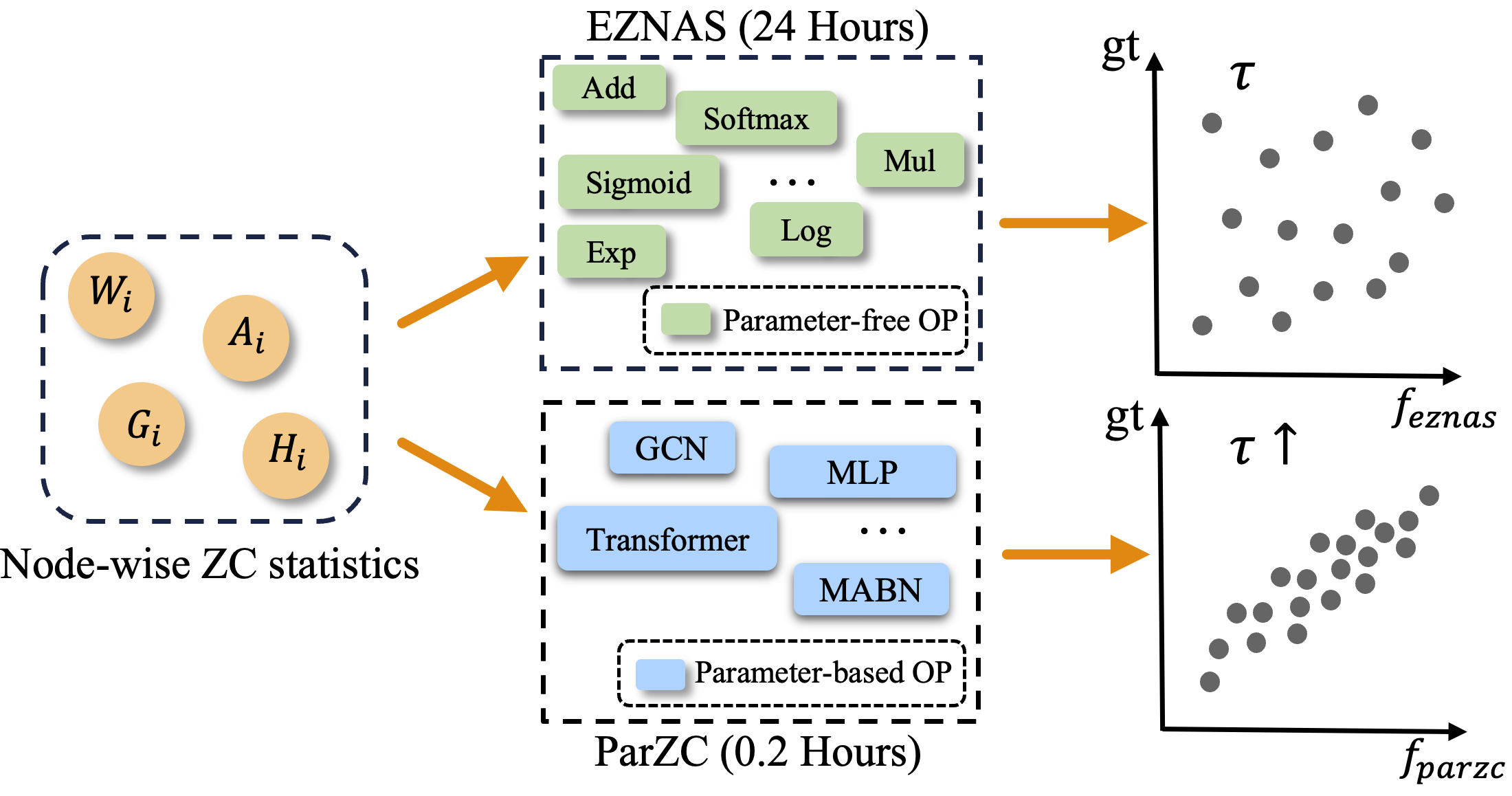}
    \vspace{-5mm}
    \caption{\textbf{Overview of ParZC and EZNAS~\cite{akhauri2022eznas} pipeline}. W: Weight, A: Activation, G: Gradient, H: Hessian Matrix.}\label{fig:diff_w_eznas}
    \vspace{-3mm}
\end{figure}

Deep learning has become indispensable in computer vision and natural language processing, with neural architecture design becoming increasingly critical. However, the traditional manual design of architectures requires extensive trial-and-error and domain knowledge, which can be time-consuming and may limit the exploration of new and innovative architectures. To address this, Neural Architecture Search (NAS) ~\cite{zoph2016neural, real2018regularized} is introduced, which offers an automated solution by traversing the search space to identify superior architectures. Despite its potential, NAS has been criticized for its substantial requirements on computational resources. For instance, NASNet~\cite{zoph2016neural} necessitated 2,000 GPU hours to identify an architecture. This significant demand for resources has impeded the broader application of NAS in practical settings. To tackle this, surrogate-based methods~\cite{chen2019pdarts, nasnet}, one-shot NAS~\cite{pham2018efficient, darts, cai2019once}, and zero-shot NAS~\cite{mellor2021neural, ZenNAS, abdelfattah2021zerocost} are investigated to expedite the process. Zero-shot NAS has attracted considerable interest with Zero-Cost (ZC) proxies. ZC proxies enable rapid scoring and ranking of untrained neural architectures based on model statistics and operations such as weight, network activation, gradient, and Hessian matrix, offering a promising avenue for reducing the computational demands of NAS.

\begin{figure}[t]
    \centering
    \includegraphics[width=1\linewidth]{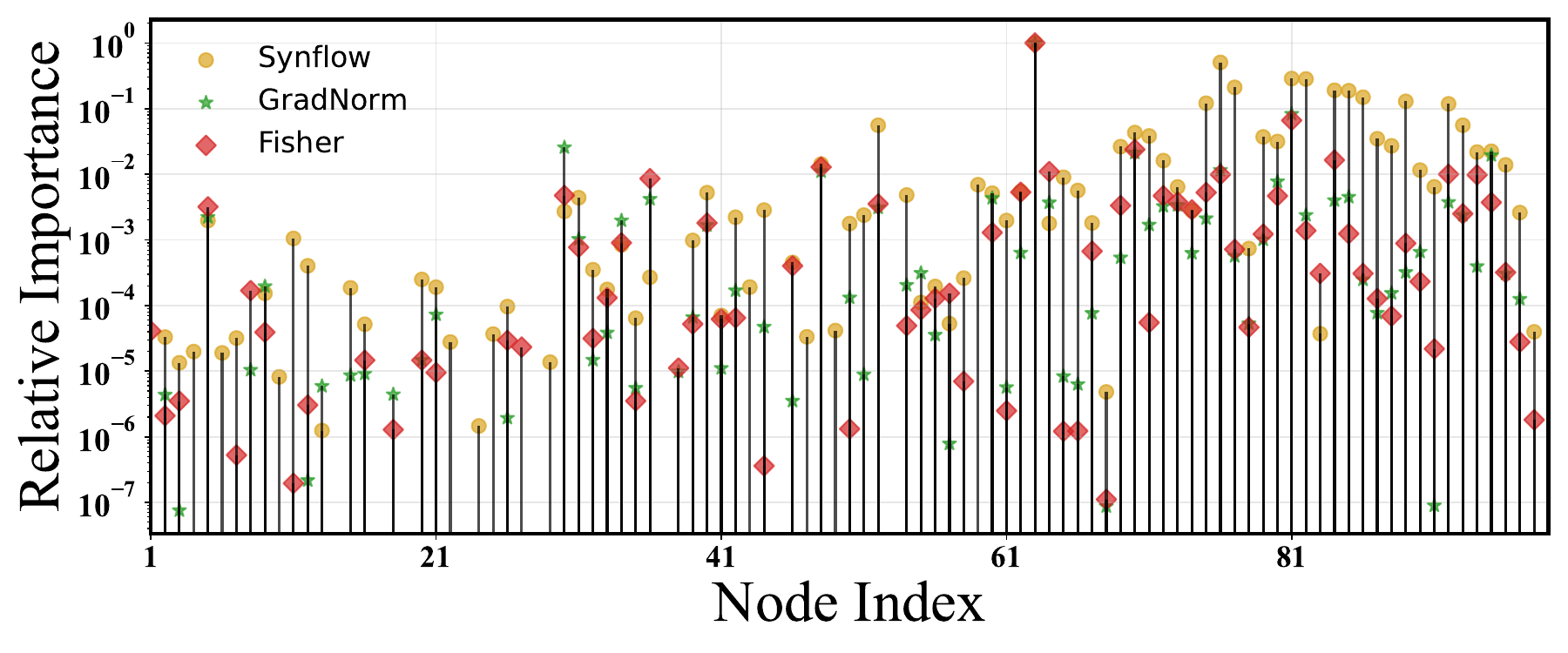}
    \vspace{-8mm}
    \caption{\textbf{Node-wise relative importance of ZC proxies} (Synflow~\cite{tanaka2020pruning_synflow}, GradNorm~\cite{abdelfattah2021zerocost}, and Fisher~\cite{Turner2019BlockSwapFB_fisher}) based on GBDT impurity on NAS-Bench-201.}\label{fig:insight}
    \vspace{-7mm}
\end{figure}

Despite its high efficiency, ZC proxy exhibits limitations, particularly in accurately ranking the top-performing architectures and its adaptability across diverse tasks. These challenges, highlighted in influential studies, underscore the imperative for further refinement in this domain. In particular, these limitations are summarized as follows:
\textbf{(1) Tedious Expert Design:} The design of ZC proxies often requires extensive expert involvement~\cite{shu2022nasi, knas, chen2020tenas} or time-consuming search processes (e.g., EZNAS~\cite{akhauri2022eznas} requires 24 hours to find a suitable proxy). Furthermore, these manually crafted ZC proxies can be susceptible to human biases (e.g., the search space of EZNAS is heavily influenced by existing ZC proxies, resulting in proxies that are similar to Synflow~\cite{tanaka2020pruning_synflow} and NWOT~\cite{mellor2021neural}).
\textbf{(2) Non-adaptive:} 
Hand-crafted ZC proxies are tailored for specific architectures, in contrast to unseen ones.  
As proved by TF-TAS~\cite{DSS}, ZC proxies tailored for CNN search space have deteriorated performance on Vision Transformer search space. 
\textbf{(3) Ranking Instability:} 
The performance of ZC proxies is notably impacted by variations in initialization methods, seed settings, and batch size, leading to inconsistent results. 
For example, DisWOT~\cite{dong2023diswot} finds the ZC proxies are sensitive to initialization methods. Additionally, under varying seeds and a batch size of one, the EZNAS revealed significant variance in ZC-NASM~\cite{akhauri2022eznas}. These observations underscore the inherent uncertainty associated with ZC proxies.
\textbf{(4) Homogeneity Assumption:} Previous ZC proxies~\cite{ mellor2021neural, ZenNAS, tanaka2020pruning_synflow, Lee2018SNIPSN, abdelfattah2021zerocost} rely on the underlying assumption that each node has an equal influence on the ZC proxy calculation, which has been questioned by subsequent works~\cite{cavagnero2022freerea,PreNAS,shu2022unifying}. Specifically, PreNAS~\cite{PreNAS} reveals that discrimination between different nodes significantly biases the performance of existing proxies~\cite{Lee2018SNIPSN, tanaka2020pruning_synflow, abdelfattah2021zerocost}. 
FreeREA~\cite{cavagnero2022freerea} leverages the magnitude of weight and gradient in Synflow by scaling them to get LogSynflow. These investigations collectively reveal a fundamental uncertainty within ZC proxies.
% From these works, we find the uncertainty lies in the ZC proxies, which is further proven by our experiment. 

These challenges have recently gained significant attention in the community. 
For example, EZNAS~\cite{akhauri2022eznas} follows the AutoML-Zero~\cite{real2020automlzero} framework, aiming to search for better ZC proxies from scratch on NAS benchmarks. EMQ~\cite{dong2023emq} introduces an evolutionary framework for discovering ZC proxies for mixed-precision quantization with an expressive search space. We provide an overview of the automatic ZC proxy designing pipeline in Figure~\ref{fig:diff_w_eznas}. These methods start by employing node-wise model statistics as input and constructing a comprehensive search space for potential proxy candidates, which includes parameter-free operations like addition, subtraction, logarithmic, and exponential functions. Subsequently, an evaluation is conducted to assess the rank correlation between predicted scores and ground truth targets.
However, the automated methods necessitate a trial-and-error approach to assess the ZC proxies within the search space, resulting in a time-consuming process (e.g., EZNAS requires 24 hours to identify a ZC proxy).
In addition, discovered proxies from these methods only achieve marginal performance gains, limited by the parameter-free search space (e.g., EZNAS achieves only 1-7\% Spearman correlation increase over NWOT~\cite{mellor2021neural}).

These limitations encourage us to revisit the essential principle of ZC proxy design: the establishment of the mapping from node-wise model statistics to the ground truth performance. 
Hand-crafted methods endeavor to approximate this mapping using a static formulation devised by experts, whereas existing automated methods necessitate iterative evaluations of proxies within the search space. Nevertheless, both hand-crafted and automated are fixed and unscalable, limiting their fitting capabilities. Parameter-based operations present a solution with greater adaptability, increased candidate diversity, and the potential for more effective mappings than their parameter-free counterparts.
An inspiration naturally arises: Can we substantially enhance proxy design by incorporating trainable parameter operations to fit the above mapping?

In this paper, we challenge the homogeneity assumption through an intuitive experiment. Initially, we compute and encode node-wise ZC statistics and actual performance of architectures in NAS-Bench-201, including Synflow~\cite{tanaka2020pruning_synflow}, GradNorm~\cite{abdelfattah2021zerocost} and Fisher~\cite{Turner2019BlockSwapFB_fisher}, following the methodology in Sec.~\ref{sec:zc_encoding}. We then employ Gradient Boosting Decision Trees (GDBT) for regression analysis, constructing an additive model forward node-wise (see supplementary for GBDT details). Subsequently, we analyze and visualize the relative importance of each node-wise ZC statistics using GDBT impurity, as depicted in Figure~\ref{fig:insight}. Notably, nodes in the deeper layer are generally more significant than shallower ones, which may exhibit minimal or no importance. Our findings reveal that node-wise ZC statistics significantly vary in their contributions to performance. These insights affirm the necessity of distinct treatment for varying ZC proxies, thereby highlighting the inherent uncertainties associated with node-wise ZC proxies. 

In light of the above analysis, we introduce the Parametric Zero-Cost Proxies (ParZC) framework, as shown in Figure~\ref{fig:diff_w_eznas}. This framework can augment the efficacy of ZC proxies with parametric operations and find better proxies in just 0.2 hours. Specifically, we propose Mixer Architecture with Bayesian Network (MABN) to learn how to rank architectures in the search space. Mixer architecture facilitates complex interactions with transformation to the input by utilizing a segment mixer. We further incorporate the Bayesian Network to assess the uncertainties within node-wise ZC statistics.
Moreover, we identify that the main focus in zero-shot NAS is achieving ranking consistency rather than precise performance estimation. In contrast to MSE loss based methods, we propose to directly optimize rank correlation by relaxing Kendall's Tau so that ParZC can effectively handle discrepancies in the ranking of architectures. Comprehensive evaluations on NAS-Bench-101~\cite{ying2019bench}, NAS-Bench-201~\cite{dong2019NASBench201}, and NDS~\cite{radosavovic2019network} benchmarks illustrate the superior performance of our ParZC than existing ZC proxies. ParZC significantly enhances both the rank correlation and the efficiency of the search process. Concurrently, we extend the application of ParZC to other domains, a.k.a. Vision Transformer (ViT) search spaces, to assess its generalizability and adaptability. 
Our key contributions can be summarized as follows:
\begin{itemize}
    \item We introduce Parametric Zero-Cost Proxies (ParZC), an adaptable ZC proxy framework that better leverages the uncertainty inherent in node-wise ZC proxies.
    \item We incorporate the Mixer Architecture with Bayesian Network (MABN) to estimate uncertainty for node-wise ZC statistics. Additionally, we introduce DiffKendall, a novel approach designed to enhance ranking capabilities.
    \item We validate ParZC's superiority through comprehensive experiments conducted on NAS-Bench-101, 201, and NDS. Experiments on the Vision Transformer search space affirm the adaptability of ParZC.
\end{itemize}

\section{Related Work}

Neural Architecture Search (NAS) endeavors to discover the optimal architecture by building different architectural designs into the search space and employing different search algorithms (e.g., reinforcement learning and evolutionary algorithm). Vanilla NAS methods~\cite{zoph2016neural, real2018regularized} need large computation budgets (e.g., 800 GPU-days) to train various candidates individually. Therefore, one-shot NAS~\cite{pham2018efficient} introduces the supernet that encompasses all possible architectures in the search space so that all architectures share the same weight and thus accelerate the convergence of candidate architectures, drastically reducing the computational resources required. In addition, many advanced NAS benchmarks~\cite{ying2019bench,dong2019NASBench201, radosavovic2019network} are built with the ground truth of a given search space. Based on these benchmarks, many predictor-based NAS methods~\cite{zela2022surrogate, wen2020neural} have been developed to bridge the input architectures and accuracy results. Recently, training-free NAS~\cite{ZenNAS,mellor2021neural,dong2023diswot}, also called zero-shot NAS, eliminated the requirements for training candidate architectures during the search phase. Zero-shot NAS employs Zero-Cost (ZC) proxies as predictive indicators to approximate the potential of architectures. This approach involves evaluating architectures using ZC proxies on randomly initialized weights, requiring a limited number of forward and backward passes with a mini-batch of input data, thereby significantly enhancing efficiency.

We view a neural network as a Directed Acyclic Graph (DAG) comprising numerous nodes, each symbolizing a specific operation. Zero-shot NAS can be categorized into two main types~\cite{ning2021evaluating} based on how to handle the neural network. (1) \textbf{Node-level zero-shot NAS} adopt from pruning literature including GradNorm~\cite{abdelfattah2021zerocost}, Plain~\cite{NIPS1988_07e1cd7d_plain}, SNIP~\cite{Lee2018SNIPSN},GraSP~\cite{Wang2020PickingWT_GraSP} Fisher~\cite{Turner2019BlockSwapFB_fisher}, and Synflow~\cite{tanaka2020pruning_synflow}. These ZC proxies are named after sensitivity indicators initially designed for fine-grained network pruning that measure the approximate loss change when certain parameters or activations are pruned. ZCNAS~\cite{abdelfattah2021zerocost} proposes to sum up node-wise sensitivities of all nodes to evaluate an architecture. (2) \textbf{Architecture-level zero-shot NAS} holistically assesses the architecture's discriminability by discerning variances among distinct input images. NWOT~\cite{mellor2021neural} proposes a heuristic metric based on local Jacobian values to estimate the performance. ZenNAS~\cite{ZenNAS} evaluates the candidate architectures using the gradient norm of the input image as a ranking score. KNAS~\cite{knas} utilizes the mean of the Gram matrix of gradients for estimation. NASI~\cite{shu2022nasi} employs the Neural Tangent Kernel~\cite{jacot2018neural_ntk} to derive an estimator based on the trace norm of the NTK matrix.

However, these methods necessitate expert design, and their formulation remains static, lacking adaptability to various tasks. Therefore, some hybrid approaches~\cite{akhauri2022eznas} based on AutoML-Zero~\cite{real2020automlzero} automatically search for the better proxies based on the ground truth from the benchmarks. These proxy search algorithms typically require a tedious proxy evolution process (24 hours) because there are a large quantities of invalid zero-cost proxy candidates due to the illegal computation or tensor shape mismatching. Our ParZC employs parameter-based operations to replace parameter-free operations, substantially reducing proxy optimization overhead and addressing the limitations of existing proxies, thereby significantly enhancing rank correlation. Diverging from training-based NAS and predictor-based NAS, our ParZC opens a new avenue for the exploration of hybrid training-free NAS approaches.

\begin{figure*}[t]
    \centering
    \includegraphics[width=1\linewidth]{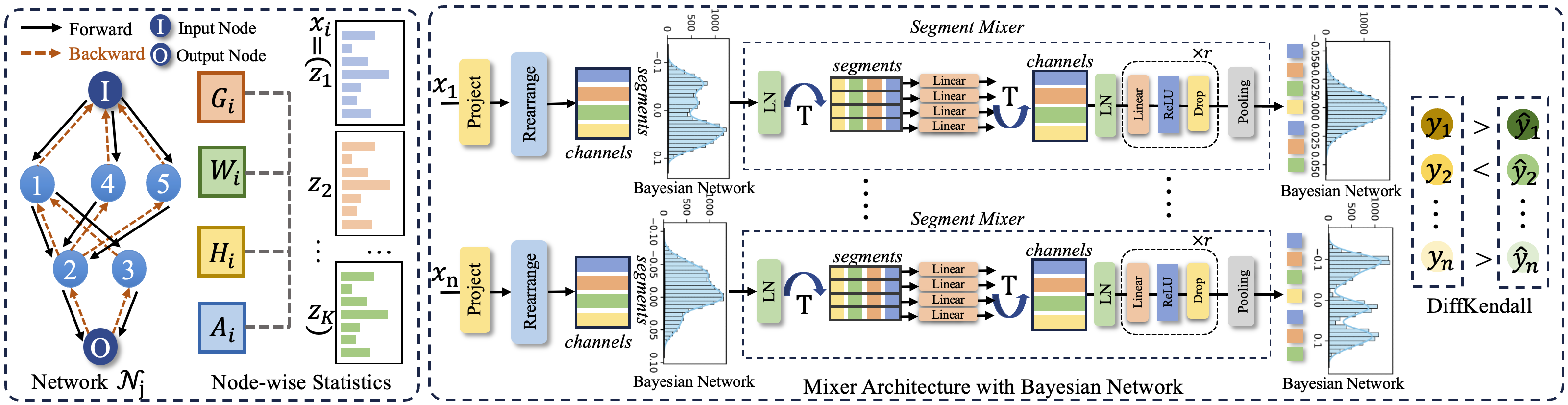}
    % \caption{Illustration of ParZC: Modules}
    \caption{\textbf{The framework of ParZC.} Left: Illustration of node-wise ZC proxies. Different ZC may extract gradient (G), weight (W), hessian (H), or activation (A) from different nodes. ParZC utilizes these node-wise ZC from different proxies as input. Right: mixer architecture with Bayesian network. We propose a Bayesian network and mixer architecture to build the ParZC to measure the uncertainty and enhance inter-channel information extraction. We propose DiffKendall as a loss function to better monitor the relative relation of different architectures.}
    \label{fig:main_figure}
    \vspace{-3mm}
\end{figure*}
\section{Parametric Zero-Cost Proxies} 
% In this section, we present parametric ZC proxies. The framework of our proposed ParZC is illustrated in Figure~\ref{fig:main_figure}, which includes the node-wise ZC aggregation process, knowledge extraction from node-wise ZC, and how to measure the uncertainty of ZC statistics with Bayesian Network. 

\subsection{Preliminary}
We present $L$-node neural network $\mathcal{N}=\{N_1,...,N_L\}$ as Directed Acyclic Graph (DAG) with weights $\mathcal{W}=\{W_1,...,W_L\}$ as shown in Figure~\ref{fig:main_figure}, where $W_i$ is the weight of the $i$-th node of $\mathcal{N}$ and each node represents operations such as Conv$1\times1$, Conv$3\times3$. For each node $N_i$, we can get the detailed statistics set $\mathcal{S}(N_i):=\{W_i, G_i, A_i, H_i\}$ where $W_i$, $G_i$, $A_i$, $H_i$ denotes the weight, gradient, activation, and Hessian matrix. The $k$-th node-wise ZC proxy, utilizing the combination of statistics, can be represented as: $z_k:\mathcal{S}(N_i) \rightarrow \mathbb{R}$. For simplicity, we denote $ z_k(\mathcal{N}):=\left\{z_k\left(\mathcal{S}(N_1)\right), \ldots, z_k(\mathcal{S}(N_L))\right\}$. We have $K$ node-wise ZC proxies $\mathcal{Z}=\{z_1,\ldots,z_k,\dots,z_K\}$. Given the $m$-th neural network $\mathcal{N}^{(m)}$ and the $k$-th ZC proxy $z_k$, we perform depth-first search (DFS) to gather node-level ZC statistics. As illustrated in Figure~\ref{fig:main_figure}, just with one batch of data as input to perform forward and backward operations, we can gather the statistic $z_k(\mathcal{N}^{(m)})\in \mathbb{R}^{L}$, where the label on the nodes denotes the order of processing. For $K$ node-wise ZC proxies on $\mathcal{N}_j^{(m)}$, we have input $x_j=\{z_1(\mathcal{N}), ..., z_K(\mathcal{N})\}\in \mathbb{R}^{(K\times L)}$ and ground truth target $y_j\in \mathbb{R}$. Therefore, we build a dataset for ZC proxies $\mathcal{D}=\{X,Y\}=\{(x_j, y_j)\}_{j=1}^{n}, X\in \mathbb{R}^{n\times (K\times L)}, Y\in \mathbb{R}^{n}$. We denote the weight of the ParZC model as $\mathcal{M}$. The estimated performance is given by $\hat{y}_n = f(x_n; \mathcal{M})$, where $f(\cdot; \mathcal{M})$ represents the output function of the ParZC model. 
We expect a high Kendall's Tau correlation between the estimated values $\hat{Y}$ and the actual values $Y$.

\subsection{Node-wise ZC Encoding}\label{sec:zc_encoding}
Due to the significant magnitude differences and substantial variations among node-wise ZC statistics from different proxies, we employ min-max feature scaling $\sigma$ as an encoding technique. This approach can mitigate the larger condition number issue and reduce variance stemming from disparate feature scales. The encoding for each ZC statistics $z_k(\mathcal{N}^{(m)})$ is defined by the equation:
\begin{equation}
    \sigma(z_k(\mathcal{N}^{(m)})):= \frac{z_k(\mathcal{N}^{(m)}) - \min(z_k(\mathcal{N}^{(m)}))}{\max(z_k(\mathcal{N}^{(m)})) - \min(z_k(\mathcal{N}^{(m)}))}
    \nonumber
\end{equation}
The encoding methods aim to normalize the feature scales and thus reduce variance. Notably, most methods, including NP~\cite{wen2020neural} and our ParZC, can only converge with this encoding technique.

\subsection{Mixer Architecture with Bayesian Network}
Stacked Multi-layer perceptions (MLPs) can approximate complex nonlinear functions but struggle in capturing intricate, higher-order interactions among input data with high uncertainty. We introduce a novel approach, the Mixer Architecture with Bayesian Network (MABN), as shown in Figure~\ref{fig:main_figure}, designed to explicitly model uncertainty by embedding probabilistic relationships within ZC proxy statistics. Given the inherent instability in ZC estimations, our method utilizes the Mixer Architecture to explore inter-segment relationships effectively. Additionally, we enhance this architecture by incorporating Bayesian networks, significantly improving its capability to assess uncertainty in node-wise ZC proxies. 

\noindent\textbf{Bayesian Network.} We introduce a Bayesian Network that employs probabilistic backpropagation~\cite{hernandez2015probabilistic}, which can enhance the estimation of the input uncertainty. 
Each Bayesian layer transforms the input $x \in \mathbb{R}^{N\times L\times P}$ to an output $y \in \mathbb{R}^{N\times L\times O}$ through a linear transformation using Bayesian weights $y = xW_{b}^T$. The weight matrix $W_b$ is computed using the reparameterization trick:
\begin{equation}
W_b = \mu + \log(1 + e^\rho) \cdot \epsilon
\nonumber
\end{equation}
where $\mu$ represents the mean and $\rho$ represents the log distribution variance, and $\epsilon \sim \mathcal{N}(0, I)$ is a random variable sampled from a standard normal distribution. In the Bayesian network, the output $Y$ given an input $X$ and weights $W_b$ is described by the conditional probability $P(Y|X, W_b)$, indicating the probability of observing $Y$ for specified $X$ and $W_b$. The weights are derived from a posterior distribution $P(W_b|X, Y)$. The process culminates in a probabilistic linear transformation $Y = XW_b^T$, where each forward pass involves integrating over potential linear transformations, weighted according to their posterior probabilities. This method aligns with Bayesian principles, effectively allowing the network to incorporate uncertainty into its predictions. As illustrated in Figure~\ref{fig:main_figure}, we incorporate Bayesian Networks both before and after the Mixer Architecture to enhance uncertainty modeling and improve the estimation of node-wise ZC proxies.

\noindent\textbf{Mixer Architecture.} We first apply a linear layer to project the input into a higher-dimensional space $X'$. We further segment $X'\in \mathbb{R}^{N\times (S\times L)}$ by splitting input into $S$ segments with length of $L$ then we have $X'\in \mathbb{R}^{N\times S\times L}$. In response to improved estimations of ZC proxies, we introduce the Mixer architecture, a concise approach designed to model the complex and nonlinear mapping of node-wise ZC statistics by exploiting inter-segment relationships. The Mixer architecture leverages a segment mixer to achieve these goals. The process begins with a preprocessing phase where a Bayesian Network (BN) assesses segment uncertainty $X_b=XW_b^T$. A layer normalization step follows BN, expressed as $X'_b = \text{LayerNorm}(X_b)$, where \(X_b\) denotes the transposed input segments. Subsequently, a Feedforward Network (FFN) is applied to these normalized segments, enhancing cross-segment interaction. This operation is represented as $X_{\text{seg}} = X'^T_{b} + \text{FFN}\left(X'^T_{b}\right)$. The segment \(X_{\text{seg}}\) is then transposed once more and processed through \(r\) combination of Linear, ReLU, and Dropout. Following a pooling operation, the segment dimensions are transformed from \( \mathbb{R}^{N\times S\times L} \) to \( \mathbb{R}^{N\times S} \). A Bayesian Network processes the output in the final stage, enabling precise estimation of the architecture's characteristics.

This methodology within each mixer block significantly bolsters the model's ability to discern and interpret complex inter-segment relations and patterns within the input data. The Bayesian MLP Mixer architecture represents a sophisticated blend of Bayesian inference principles and structured segment mixing. This innovative approach marks a leap forward from traditional MLP architectures, offering enhanced capabilities in processing and understanding intricate data structures. Our structure exhibits a resemblance to that of MLP-Mixer~\cite{mixer}. However, a notable disparity lies in our input methodology. Unlike MLP-Mixer, which splits images into multiple patches, our mixer architecture exclusively relies on probability as its input source.

\subsection{Differentiable Ranking Optimization}
We employ Kendall's tau to assess the correlation between the rankings produced by zero-shot estimations and the ground truth. However, the standard form of Kendall's tau is not differentiable, complicating its use in gradient-based optimization. To make Kendall's tau differentiable, we proposed DiffKendall, which introduces a sigmoid-based transformation characterized by parameters $\alpha$, encapsulated in the function $\sigma_{\alpha}(\Delta) = \text{sigmoid}(\alpha \Delta) - \text{sigmoid}(-\alpha \Delta)$. This transformation smooths the non-differentiable sign function inherent in the original Kendall's Tau computation. The approximation of Kendall's Tau $\tau_d$, is then articulated as:
\begin{equation}
\tau_d = -\frac{1}{\binom{L}{2}} \sum_{i \neq j} \sigma_{\alpha}(\Delta x_{ij}) \cdot \sigma_{\alpha}(\Delta y_{ij})
\nonumber
\end{equation}
where ${\binom{L}{2}}$ represents the total number of unique element pairs and $\Delta x_{ij}=x_i-x_j, \Delta y_{ij}=y_i-y_j$. This expression encapsulates the concordance and discordance between the ranks of elements in the sequences $x$ and $y$ while maintaining differentiability.

In contrast to the pairwise rank loss~\cite{xu2021renas}, which relies on the quality of the pairs selected for training, the proposed $\tau_{d}$ offers a broader view of rank correlation by considering both concordant and discordant pairs in sequences. This holistic approach provides a more nuanced understanding of rank relationships, especially in contexts where global rank correlation is crucial. Thus, $\tau_{d}$ can be a compelling alternative, particularly when capturing global rank correlation is essential.

\begin{table*}[t]
\centering
\caption{\textbf{Spearman (SP) and Kendall's Tau (KD) correlation coefficients (\%) of various ZC proxies} across NAS benchmarks NAS-Bench-101 (NB101), NAS-Bench-201 (NB201), and NDS for CIFAR-10, CIFAR-100, and ImageNet16-120 datasets.}\label{tab:nb101_201_nds}
\resizebox{\textwidth}{!}{
\begin{tabular}{l|cccccccccccccc}
\toprule
         & \multicolumn{2}{c}{NB101-CF10}                              & \multicolumn{2}{c}{NB201-CF10}                              & \multicolumn{2}{c}{NB201-CF100}                             & \multicolumn{2}{c}{NB201-IMG16}                             & \multicolumn{2}{c}{NDS-DARTS}                               & \multicolumn{2}{c}{NDS-NASNet}                              & \multicolumn{2}{c}{NDS-ENAS}                                \\ \cline{2-15} 
         & \multicolumn{1}{c}{SP}       & \multicolumn{1}{c}{KD}       & \multicolumn{1}{c}{SP}       & \multicolumn{1}{c}{KD}       & \multicolumn{1}{c}{SP}       & \multicolumn{1}{c}{KD}       & \multicolumn{1}{c}{SP}       & \multicolumn{1}{c}{KD}       & \multicolumn{1}{c}{SP}       & \multicolumn{1}{c}{KD}       & \multicolumn{1}{c}{SP}       & \multicolumn{1}{c}{KD}       & \multicolumn{1}{c}{SP}       & \multicolumn{1}{c}{KD}       \\ \midrule
Params   & 37.0  & 25.0  & 72.0 & 54.0 & 73.0  & 55.0 & 69.0 & 52.0 & 67.0  & 50.0  & 50.5 & 36.1 & 41.0 & 32.0 \\
FLOPs    & 36.0  & 25.0  & 69.0 & 50.0 & 71.0  & 52.0 & 67.0 & 48.0 & 67.6  & 50.7  & 48.1 & 34.5 & 41.0 & 32.0 \\ \midrule
Fisher~\cite{Turner2019BlockSwapFB_fisher}   & -28.0 & -20.0 & 50.0 & 37.0 & 54.0  & 40.0 & 48.0 & 36.0 & 33.7  & 22.7  & -9.2 & -4.8 & -5.9 & -4.1 \\
GradNorm~\cite{abdelfattah2021zerocost} & -25.0 & -17.0 & 58.0 & 42.0 & -63.0 & 47.0 & 57.0 & 42.0 & 37.5  & 26.0  & -7.1 & -3.9 & -0.4 & -0.1 \\
GraSP~\cite{Wang2020PickingWT_GraSP}    & 27.0  & 18.0  & 51.0 & 35.0 & 54.0  & 38.0 & 55.0 & 39.0 & -20.8 & -14.7 & 14.2 & 8.6  & 18.4 & 12.3 \\
L2Norm~\cite{abdelfattah2021zerocost}   & 50.0  & 35.0  & 68.0 & 49.0 & 72.0  & 52.0 & 69.0 & 50.0 & 51.9  & 38.4  & 22.4 & 16.4 & 21.3 & 15.9 \\
SNIP~\cite{Lee2018SNIPSN}     & -19.0 & -14.0 & 58.0 & 43.0 & -63.0 & 47.0 & 57.0 & 42.0 & 42.3  & 30.0  & -0.7 & 0.9  & 2.8  & 2.6  \\
Synflow~\cite{tanaka2020pruning_synflow}  & 31.0  & 21.0  & 73.0 & 54.0 & 76.0  & 57.0 & 75.0 & 56.0 & 49.9  & 36.4  & 7.5  & 5.3  & 6.3  & 4.0  \\ \midrule
NWOT~\cite{mellor2021neural}     & 31.0  & 21.0  & 77.0 & 58.0 & 80.0  & 62.0 & 77.0 & 59.0 & 66.3  & 48.9  & 44.9 & 31.7 & 38.0 & 28.0 \\
Zen~\cite{ZenNAS}      & 59.0  & 42.0  & 35.0 & 27.0 & 35.0  & 28.0 & 39.0 & 29.0 & 49.0  & 36.1  & 13.2 & 10.2 & 13.5 & 10.4 \\
ZiCo~\cite{li2023zico}     & 63.0  & 46.0  & 74.0 & 54.0 & 78.0  & 58.0 & 79.0 & 60.0 & 49.5  & 34.9  & 22.4 & 16.7 & 17.3 & 12.0 \\
EZNAS~\cite{akhauri2022eznas}    & 6.8   & 4.5   & 83.0 & 65.0 & 82.0  & 65.0 & 78.0 & 61.0 & 67.0  & \textbf{56.0}  & 50.0 & 44.0 & 63.0 & 52.0 \\
\rowcolor{gray!30} ParZC    & \textbf{83.2}  & \textbf{63.7}  & \textbf{90.4} & \textbf{70.6} & \textbf{91.1}  & \textbf{74.3} & \textbf{87.9} & \textbf{69.9} & \textbf{67.8}  & 50.3  & \textbf{54.9} & \textbf{38.5} & \textbf{69.0} & \textbf{50.6} \\ \bottomrule
\end{tabular}}\label{tab:nb201_acc}
\end{table*}

\begin{table}[ht]
    \centering
    % \caption{Ranking results on NAS-Bench-101 and NAS-Bench-201.}
    \caption{\textbf{Kendall's Tau Coefficients (\%) for Training-Based NAS Algorithms} on CIFAR-10, evaluated on NAS-Bench-101 and NAS-Bench-201, illustrating ranking performance across subsets with varying sample sizes. $^\dagger$: CTNAS~\cite{chen2021contrastive}, $^\ddagger$: TNASP~\cite{lu2021tnasp}, $^*$: PINAT~\cite{PINATAAAI23}.}\label{tab:predictor_nb101_nb201}
    \resizebox{80mm}{!}{
    \begin{tabular}{lccccc}
        \toprule
        NAS-Bench-101 & $S_{100}$ & $S_{172}$ & $S_{424}$ & $S_{424}$ & $S_{4236}$ \\
        \midrule
        SPOS~\cite{guo2020single}$^\dagger$ & - & - & 19.6 & - & - \\
        FairNAS~\cite{chu2019fairnas}$^\dagger$ & - & - & 23.2 & - & - \\ 
        ReNAS~\cite{xu2021renas}$^\dagger$ & - & - & 63.4 & 65.7 & 81.6 \\
        %RegressionNAS$^\dagger$ & - & - & 0.430 & - & - \\
        NP~\cite{wen2020neural}$^\ddagger$ & 39.1 & 54.5 & 71.0 & 67.9 & 76.9 \\
        NAO~\cite{luo2018neural}$^\ddagger$ & 50.1 & 56.6 & 70.4 & 66.6 & 77.5 \\
        Arch2Vec~\cite{yan2020Arch2Vec}$^*$ & 43.5 & 51.1 & 56.1 & 54.7 & 59.6 \\
        %D-VAE$^*$ & 0.530 & 0.549 & 0.671 & 0.626 & 0.698 \\
        GATES~\cite{ning2020GATES}$^*$ & 60.5 & 65.9 & 66.6 & 69.1 & 82.2 \\
        %GraphTrans$^*$ & 0.330 & 0.472 & 0.600 & 0.602 & 0.700 \\
        %Graphormer$^*$ & 0.564 & 0.580 & 0.596 & 0.611 & 0.797 \\
        CTNAS~\cite{chen2021contrastive}$^\dagger$ & - & - & 75.1 & - & - \\
        TNASP~\cite{lu2021tnasp}$^\ddagger$ & 60.0 & 66.9 &75.2 & 70.5 & 82.0 \\
        %GMAE-NAS$^*$ & 0.666 & 0.697 & 0.788 & 0.732 & 0.775 \\
        PINAT~\cite{PINATAAAI23}$^*$ & 67.9 & 71.5 & \textbf{80.1} & 77.2 & 84.6 \\
        %\midrule 
    \rowcolor{gray!30}     \textbf{ParZC} & \textbf{69.3} & \textbf{71.7} & 79.7 & \textbf{78.2} & \textbf{85.3} \\
        \midrule
        NAS-Bench-201 & $S'_{78}$ & $S'_{156}$ & $S'_{469}$ & $S'_{781}$ & $S'_{1563}$ \\
        \midrule
        NP~\cite{wen2020neural}$^\ddagger$ & 34.3 & 41.3 & 58.4 & 63.4 & 64.6 \\
        NAO~\cite{luo2018neural}$^\ddagger$ & 46.7 & 49.3 & 47.0 & 52.2 & 52.6 \\
        Arch2Vec~\cite{yan2020Arch2Vec}$^*$ & 54.2 & 57.3 & 60.1 & 60.6 & 60.5 \\
        %GraphTrans$^*$ & 0.409 & 0.550 & 0.594 & 0.588 & 0.673 \\
        %Graphormer$^*$ & 0.505 & 0.630 & 0.680 & 0.719 & 0.776 \\
        TNASP~\cite{lu2021tnasp}$^\ddagger$ & 53.9 & 58.9 & 64.0 & 68.9 & 72.4 \\
        PINAT~\cite{PINATAAAI23}$^*$ & 54.9 & 63.1 & 70.6 & 76.1 & 78.4 \\
       % \midrule 
      \rowcolor{gray!30}   \textbf{ParZC} & \textbf{64.6} & \textbf{70.6} & \textbf{80.6} & \textbf{83.2} & \textbf{85.5} \\
        \bottomrule
    \end{tabular}
     }
    %\footnotesize
    %\newline
    
\end{table}

\begin{table*}[ht]
\caption{\textbf{Comparison of NAS algorithms on NAS-Bench-201}. The result of ParZC is reported with mean and standard deviation of 3 independent runs. ``C" and ``D" denotes continuous and discrete search space.}
\label{tab:sota-nasbench201}
\centering
\resizebox{0.9\linewidth}{!}{
\begin{tabular}{lccccccc}
\toprule
\multirow{2}{*}{\textbf{Algorithm}} & \multicolumn{3}{c}{\textbf{Test Accuracy (\%)}} &
\multirow{2}{*}{\textbf{Cost}} &
\multirow{2}{*}{\textbf{Method}} &
\multirow{2}{*}{\textbf{Applicable}} \\
\cmidrule(l){2-4} 
& CIFAR-10 & CIFAR-100 & ImageNet16-120 & (GPU Sec.) & & \textbf{Space}\\
\midrule 
ResNet ~\cite{he2016resnet} & 93.97 & 70.86 & 43.63 & - & manual & - \\
\midrule
REA$^{\dagger}$ & 93.92$\pm$0.30 & 71.84$\pm$0.99 & 45.15$\pm$0.89 & 12000 & evolution & C \& D \\
RS (w/o sharing)$^{\dagger}$ & 93.70$\pm$0.36 & 71.04$\pm$1.07 & 44.57$\pm$1.25 & 12000 & random & C \& D \\
REINFORCE$^{\dagger}$ & 93.85$\pm$0.37 & 71.71$\pm$1.09 & 45.24$\pm$1.18 & 12000 & RL & C \& D \\
BOHB$^{\dagger}$ & 93.61$\pm$0.52 & 70.85$\pm$1.28 & 44.42$\pm$1.49 & 12000 & BO+bandit & C \& D \\
\midrule
ENAS$^{\ddagger}$ ~\cite{enas} & 93.76$\pm$0.00 & 71.11$\pm$0.00 & 41.44$\pm$0.00 & 15120 & RL & C \\
%DARTS (1st)$^{\ddagger}$ ~\cite{darts} & 54.30$\pm$0.00 & 15.61$\pm$0.00 & 16.32$\pm$0.00 & 16281 & gradient & C \\
%DARTS (2nd)$^{\ddagger}$ ~\cite{darts} & 54.30$\pm$0.00 & 15.61$\pm$0.00 & 16.32$\pm$0.00 & 43277 & gradient & C \\
GDAS$^{\ddagger}$ ~\cite{gdas} & 93.44$\pm$0.06 & 70.61$\pm$0.21 & 42.23$\pm$0.25 & 8640 & gradient & C \\
DrNAS$^{\sharp}$ ~\cite{chen2021drnas} & 93.98$\pm$0.58 & 72.31$\pm$1.70 & 44.02$\pm$3.24 & 14887 & gradient & C \\
\midrule
NWOT ~\cite{mellor2021neural} & 92.96$\pm$0.81 & 69.98$\pm$1.22 &  44.44$\pm$2.10 & 306 & training-free & C \& D\\
TE-NAS ~\cite{chen2020tenas} & 93.90$\pm$0.47 & 71.24$\pm$0.56 & 42.38$\pm$0.46 & 1558 & training-free & C \\
KNAS ~\cite{knas} & 93.05 & 68.91 & 34.11 & 4200 & training-free & C \& D \\
NASI ~\cite{shu2022nasi} & 93.55$\pm$0.10 & 71.20$\pm$0.14 & 44.84$\pm$1.41 & 120 & training-free & C \\
GradSign ~\cite{gradsign} & 93.31$\pm$0.47 & 70.33$\pm$1.28 & 42.42$\pm$2.81 & - & training-free & C \& D \\
\midrule
EZNAS~\cite{akhauri2022eznas} & 93.63$\pm$0.12 & 69.82$\pm$0.16 & 43.47$\pm$0.20 & - & hybrid & D \\

\rowcolor{gray!30}  ParZC& \textbf{94.36}$\pm$0.01 & \textbf{73.49}$\pm$0.02 & \textbf{46.34}$\pm$0.04 & \textbf{68} & hybrid & C \& D \\
\midrule
\textbf{Optimal} & 94.37 & 73.51 & 47.31 & - & - & -\\
\bottomrule
\end{tabular}
}
% \end{small}
\vskip -0.1in
\end{table*}
\section{Experiments}

\subsection{Datasets and Implementation Details} We conduct experiments on various NAS benchmarks with extensive search space including NAS-Bench-101 (NB101)~\cite{ying2019bench}, NAS-Bench-201 (NB201)~\cite{dong2019NASBench201} and Network Design Spaces (NDS)~\cite{radosavovic2019network} with DARTS~\cite{darts}/NASNet~\cite{zoph2016neural}/ENAS~\cite{pham2018efficient}, spanning CIFAR-10, CIFAR-100, and ImageNet16-120 datasets. To verify the adaptability of ParZC, we extend the experiment to ViT search space, a.k.a. Autoformer~\cite{chen2021autoformer}, on Imagenet-1k. Training is conducted with a training set, and we measure the ranking ability based on the validation dataset. For each architecture in the training set, we aggregate their node-wise ZC statistics with Synflow~\cite{tanaka2020pruning_synflow}, SNIP~\cite{Lee2018SNIPSN}, GradNorm~\cite{abdelfattah2021zerocost}, etc. We adopt Kendall's Tau (KD) and Spearman (SP) to measure the rank correlation between predicted and actual accuracy. For NB101 and NB201, we utilize Adam optimizer with a learning rate 1e-4 and weight decay of 1e-3. The training batch size is 10, and the evaluation batch size is 50. The training epochs on NB101, NB201, and NDS are 150, 200, and 296, respectively. Specifically for NDS, we mainly conduct experiments on NASNet, DARTS, and ENAS search spaces to verify the ranking ability of ParZC. DiffKendall is a loss function when training ParZC with $\alpha=0.5$. We detail the training settings in the supplementary for different search spaces. All of the experiments are conducted on GeForce RTX 4090Ti and PyTorch~\cite{paszke2019pytorch} framework. The hyperparameters of our proposed MABN, such as hidden size, dropout rate, and embedding dimension, are finely tuned using Bayesian optimization with Optuna~\cite{optuna_2019}. For more details, please refer to the supplementary. 

\subsection{Experiments on NAS Benchmarks} 

\noindent\textbf{Comparison with ZC Proxies.} We report the rank correlation with Spearman (SP) and Kendall's Tau (KD) on three NAS benchmarks in Table~\ref{tab:nb101_201_nds}, including NB101, NB201 and NDS. 
The results on NB101 and NB201 are obtained from previous methods~\cite{abdelfattah2021zerocost,akhauri2022eznas,2023sigeo}, while the results on NDS are evaluated by us using the official implementation.
We compare our ParZC with three kinds of zero-shot NAS methods: size-based, pruning-based, and theory-based proxies. The size-based proxies serve as the baseline, encompassing FLOPs and Params, achieving competitive performance. Pruning-based proxies are inspired by pruning metrics like Fisher~\cite{Turner2019BlockSwapFB_fisher}, GradNorm~\cite{abdelfattah2021zerocost}, GraSP~\cite{Wang2020PickingWT_GraSP}, Jacov, L2Norm\cite{abdelfattah2021zerocost}, Plain~\cite{NIPS1988_07e1cd7d_plain}, SNIP~\cite{Lee2018SNIPSN}, Synflow~\cite{tanaka2020pruning_synflow}, which also achieve relatively good performance but most of them still fail to outperform the baseline. Theory-based proxies such as NWOT~\cite{mellor2021neural}, Zen~\cite{ZenNAS}, and ZiCo~\cite{li2023zico}, generally achieve better performance than pruning-based proxies but also show poor correlation on challenging search space such as NASNet and ENAS. Overall, EZNAS~\cite{akhauri2022eznas} demonstrate its superiority for ranking ability among all search space except NB101. Our proposed ParZC surpasses the baseline by a large margin and achieves competitive results across all search spaces. 

\noindent\textbf{Comparison with Training-based NAS.} To make a fair comparison, we present the Spearman correlation on NB101 and NB201 compared to training-based NAS with the same data-splitting settings. Table~\ref{tab:predictor_nb101_nb201} presents the \textbf{Kendall's Tau} coefficients for various data splits $S_{\#samples}$ within the NB101 and $S'_{\#samples}$ for NB201 benchmark. We compare our ParZC with one-shot NAS~\cite{guo2020single, chu2019fairnas} and predictor-based NAS methods~\cite{wen2020neural, PINATAAAI23, lu2021tnasp}. Note that we incorporate the operation encoding and adjacency matrix following PINAT~\cite{PINATAAAI23} into ParZC to make a comparison. Please refer to the supplementary for more details. For NB101, results demonstrate that our proposed ParZC exhibits a remarkable ability in ranking architectures on NB101, which not only outperforms the one-shot based NAS like SPOS~\cite{guo2020single} and FairNAS~\cite{chu2019fairnas} but also surpasses the SOTA transformer-based predictors like CTNAS~\cite{chen2021contrastive}, TNASP~\cite{lu2021tnasp} and PINAT~\cite{PINATAAAI23}. For NB201, our ParZC surpasses other methods by a large margin, increasing the Spearman coefficient by around 10\%. We also find that with only 78 samples (0.05\% of search space), our ParZC can achieve better performance than PINAT~\cite{PINATAAAI23} with 156 samples, which denotes that our ParZC contains additional information over the architecture and is complementary to existing predictor-based methods.

\noindent\textbf{Search Results on NAS-Bench-201.} We present a thorough evaluation of various NAS algorithms, focusing on their performance on the test set on CIFAR-10/100 and ImageNet16-120 in NB201, as detailed in Table~\ref{tab:sota-nasbench201}. To substantiate the effectiveness and efficiency of our proposed ParZC, we conduct comparative analyses with several baseline approaches, including optimization-based\cite{dong2019NASBench201}, one-shot~\cite{enas, gdas, chen2021drnas}, zero-shot~\cite{mellor2021neural, chen2020tenas, knas, shu2022nasi, gradsign} and automatic designed proxies~\cite{akhauri2022eznas}.
We categorize the various methodologies in NAS into five distinct types: evolution, random search, reinforcement, gradient, and training-free. Hybrid denotes a combination of these types. For example, EZNAS belongs to training-free and evolution categories. Our ParZC uniquely integrates gradient and training-free approaches.
As detailed in Table~\ref{tab:sota-nasbench201}, ParZC outperforms training-based and training-free baselines by consistently selecting superior performance architectures. ParZC requires only 68 GPU seconds for its search process, as it estimates performance in batches, which is significantly shorter than even zero-shot NAS methods like GradSign~\cite{gradsign}. Furthermore, our ParZC model attains SOTA accuracy on NB201 with minimal variance, showcasing its efficiency and effectiveness.

\begin{table}[ht]
\centering
\newcommand{\tabincell}[2]{\begin{tabular}{@{}#1@{}}#2\end{tabular}}
\caption{\textbf{Comparison with Vision Transformers on Imagenet-1k}. The result of ParZC is searched in the AutoFormer search space.  
% 
% $*$ denotes the results reported by ~\cite{17}.
}
\label{tab:ImageNet}
% \vspace{-2mm}
\setlength{\tabcolsep}{2.4mm}
\small{
 \resizebox{1\linewidth}{!}{
\begin{tabular}{l|cc|c}
\toprule
% \hline 
Algorithms       & \multicolumn{1}{c}{\tabincell{c}{Param (M)}} & \tabincell{c}{Top-1 (\%)}  & GPU Days \\
\midrule
% MobileNet-V2~\cite{39} &  3.5                      & -                          & 72.0                       & -                        &    CNN     & Manual           &      -       \\
Deit-Ti~\cite{Touvron2020TrainingDEIT}     &  5.7                           &   72.2                       &       -      \\
TNT-Ti~\cite{han2021transformer_tnt}     &  6.1                           &  73.9     &       -      \\
ViT-Ti~\cite{Dosovitskiy2021AnII}     &  5.7                           & 74.5                                 &      -       \\
PVT-Tiny~\cite{Wang2021PyramidVT_pvt} &  13.2                           &  75.1                            &      -       \\
ViTAS-C~\cite{Su2021ViTASVT}    &  5.6                           &  74.7                       & 32 \\
AutoFormer-Ti~\cite{chen2021autoformer}&  5.7                           & 74.7                     & 24       \\
TF-TAS-Ti~\cite{DSS}    & 5.9                           &  75.3                        &  0.5          \\
\rowcolor{gray!30}  ParZC    &  6.1                          & \textbf{75.5}                         &  \textbf{0.05}           \\
\bottomrule
\end{tabular}
}}
\vspace{-2mm}
\end{table}

\begin{table}[htbp]
  \centering
  \small
    \caption{\textbf{Comparison with ZC proxies in the Autoformer search space.} The results are reported with mean and standard deviation of 3 runs with different seeds.}
  \resizebox{0.98\linewidth}{!}{
    \begin{tabular}{l|ccc}
    \toprule
    ZC Proxies & Kendall's Tau (\%)    &Spearman (\%)  & Pearson (\%)   \\
    \midrule
    % Grasp & -2.37$_{\pm 0.05}$ & -7.53$_{\pm 1.30}$ & -2.56$_{\pm 0.08}$ \\
    SNIP~\cite{Lee2018SNIPSN} & 14.6$\pm 1.5$ & 30.6$\pm 6.0$ & 49.4$\pm 10.6$ \\
    Synflow~\cite{tanaka2020pruning_synflow} & 14.8$\pm 2.3$  & 27.6$\pm 7.2$ & 44.2$\pm 10.3$   \\
    NWOT~\cite{mellor2021neural}  & 13.3$\pm 0.1$  & 19.7$\pm 1.5$  & 38.4$\pm 9.9$ \\
    TF-TAS~\cite{DSS}  & 14.5$\pm 1.7$ & 29.9$\pm 6.3$ & 48.7$\pm 11.0$ \\
\rowcolor{gray!30}     ParZC & \textbf{41.4$\pm 0.4$} & \textbf{65.0$\pm 1.1$} & \textbf{54.1$\pm 4.1$} \\
    \bottomrule
    \end{tabular}%
    }
  \label{tab:zc_proxies_vit}
\end{table}%

\begin{table}[t]
    \caption{\textbf{Ablation study of design choices} in ParZC using 78 samples on NB201. NP: Neural Predictor~\cite{wen2020neural}, Mixer: Mixer Architecture, BN: Bayesian Network, MLP: Multi-layer Perceptron. The results are reported with mean and std of 3 runs with different seeds.}
    \vspace{-3mm} % Adjusting space above the table
    \small % Setting the font size
    \centering % Centering the table
        \resizebox{80mm}{!}{
    \begin{tabular}{cccc|cc} % Column definitions
        \toprule
        % Table header
        NP & Mixer & BN & MLP & KD(\%) & SP(\%) \\ \midrule
        % Table rows
         $\checkmark$ & - & - & - & 34.29$\pm 0.42$ & 45.61$\pm 12.89$ \\
        - & $\checkmark$ & - & - & 62.29$\pm 3.31$ & 80.19$\pm 1.45$ \\
        - & - & - & $\checkmark$ & 54.64$\pm 8.60$ & 73.50$\pm 13.78$ \\
         - & - & $\checkmark$ & - & 50.12$\pm 5.35$ & 69.09$\pm 8.86$ \\
        $\checkmark$ & $\checkmark$ & - & - & 59.79$\pm 2.41$ & 78.84$\pm 4.61$ \\
        $\checkmark$ & - & $\checkmark$ & - & 54.81$\pm 1.22$ & 74.24$\pm 1.51$ \\
        - & $\checkmark$ & $\checkmark$ & - & 67.69$\pm 4.21$  & 86.03$\pm 3.53$\\
        $\checkmark$ & $\checkmark$ & $\checkmark$ & - & \textbf{68.89$\pm 1.40$} & \textbf{87.17$\pm 0.64$} \\ 
        \bottomrule
    \end{tabular}
    }
    \label{tab:ablation-module} % Label for referencing
    \vspace{-3mm} % Adjusting space below the table
\end{table}

\begin{table}[t]
    \caption{\textbf{Ablation study of loss functions} using 178 Samples on NB101. MSE: Mean Squared Error Loss, Rank Loss: Ranking-Based Loss Function as in ReNAS~\cite{xu2021renas}, DiffKendall: Differentiable Kendall's Tau.}
    \vspace{-3mm}
    \small 
	\centering
     \resizebox{80mm}{!}{
	\begin{tabular}{ccc|cc}
		\toprule
		MSE Loss     & Rank Loss    & DiffKendall   & KD(\%) &  SP(\%) \\ \midrule
		$\checkmark$ & -            & -             & 65.69           &  85.04 \\
		-            & $\checkmark$ & -             & 65.64           &  84.89 \\
		$\checkmark$ & $\checkmark$ & -             & 64.62           &  83.92 \\
		$\checkmark$ & -            & $\checkmark$  & 65.56           &  84.75 \\
		-            & $\checkmark$ & $\checkmark$  & 64.41           &  83.85 \\
		$\checkmark$ & $\checkmark$ & $\checkmark$  & 66.36           &  85.51 \\
		-            & -            & $\checkmark$  & \textbf{66.83}  &  \textbf{85.97} \\ \bottomrule
	\end{tabular}\label{tab:ablation_diffkd}
 }
\vspace{-3mm}
\end{table}

\begin{figure}[htbp]
  \begin{minipage}[b]{0.495\linewidth}
    \centering
    \includegraphics[width=\linewidth]{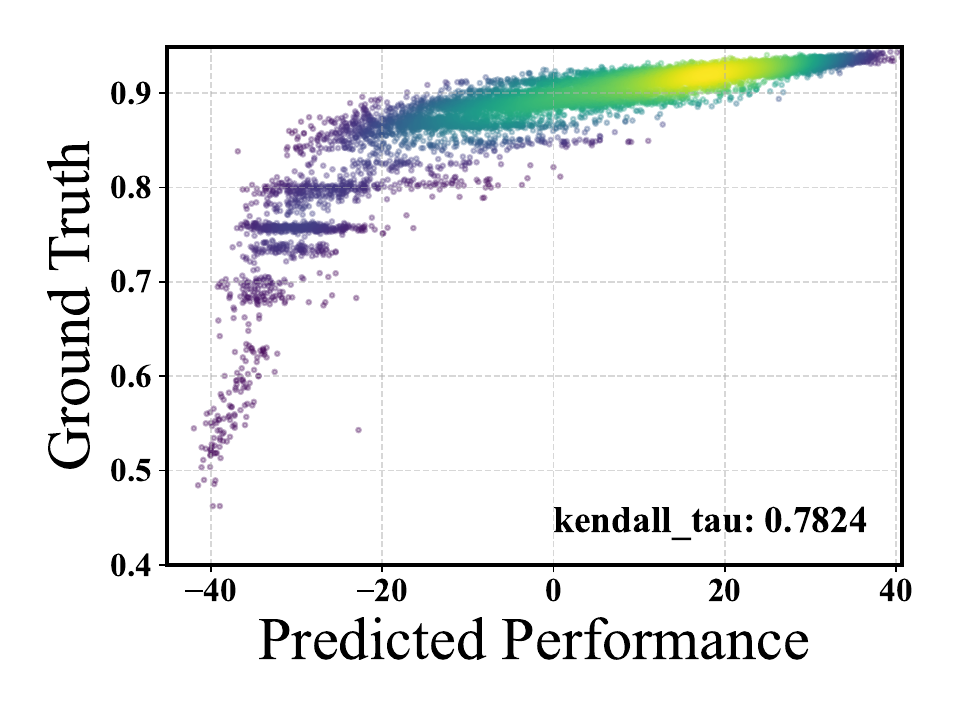}
    \subcaption{Correlation of all architectures}
    \label{fig:subfig1}
  \end{minipage}%
  \begin{minipage}[b]{0.505\linewidth}
    \centering
    \includegraphics[width=\linewidth]{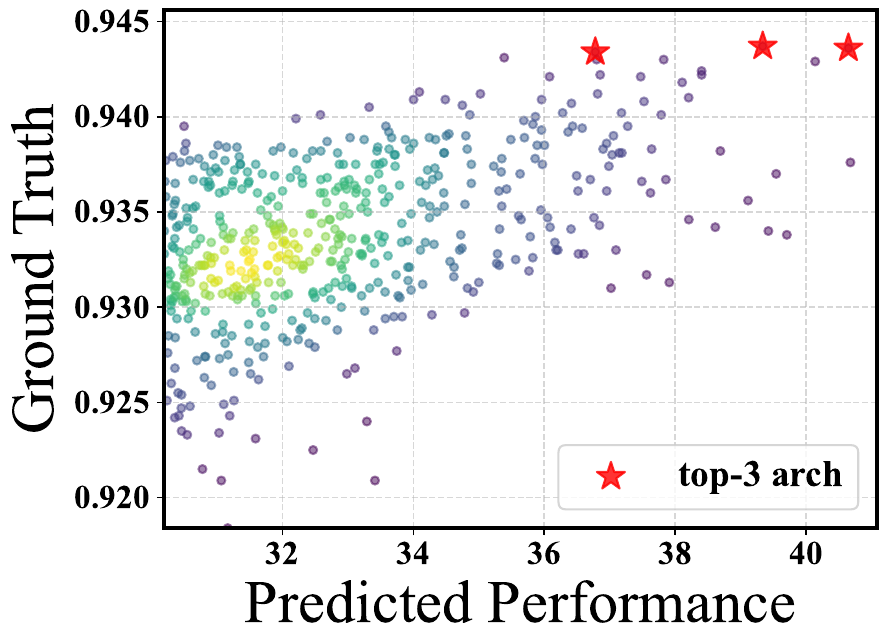}
    \subcaption{Correlation of top architectures}
    \label{fig:subfig2}
  \end{minipage}
  \caption{Rank correlation between ParZC and ground truth}
  \label{fig:vis_corr}
  \vspace{-5mm}
\end{figure}

\subsection{Experiments on Vision Transformer} 

\noindent\textbf{Search Results on Vision Transformer.} We present the performance of the searched Vision Transformer architecture on the ImageNet-1k dataset in Table~\ref{tab:ImageNet}. The AutoFormer supernet~\cite{chen2021autoformer} generates ground-truth labels, offering an efficient and cost-effective training data generation approach. Utilizing a relatively small dataset comprising 1,000 samples divided into an 80-20\% training-validation split, our ParZC algorithm demonstrates the capability to identify a high-performance architecture within an impressively short span of 0.05 GPU days. This efficiency level aligns with that of leading one-shot NAS methods. The results in the table show that the architecture identified by ParZC not only competes with but also exceeds the performance of the SOTA TF-TAS-Ti model~\cite{DSS} while maintaining a comparable number of parameters.

\noindent\textbf{Rank correlation in Vision Transformer Search Space.} To evaluate the generalization capabilities of ParZC, we assess its ranking consistency within the ViT search space, explicitly focusing on Autoformer-Ti~\cite{chen2021autoformer}. In this context, we analyze the performance of various ZC proxies. Among these, TF-TAS~\cite{DSS} is tailored explicitly for the ViT search space, while others like SNIP~\cite{Lee2018SNIPSN}, Synflow~\cite{tanaka2020pruning_synflow}, and NWOT~\cite{mellor2021neural} are initially designed for CNN search spaces. As depicted in Table~\ref{tab:zc_proxies_vit}, our ParZC algorithm achieves superior rank correlation. This result highlights ParZC's adaptability and underscores its enhanced efficiency in ranking within transformer-based search spaces.

\subsection{Ablation Study}

\noindent\textbf{Different Design Choices.} We first dissect the contributions of components in ParZC to its ranking consistency with Kendall's Tau and Spearman on NB201, as shown in Table~\ref{tab:ablation-module}. Integrating all components yields optimal KD and SP coefficients of 69.98\% and 87.90\%, respectively. Only mixer architecture is also a competitive baseline with 62.29\% KD and 80.19\% SP. Compared with baseline MLP, mixer architecture can achieve a higher rank correlation with 7.65\% higher in KD and 6.69\% higher in SP. We also observe that NP~\cite{wen2020neural} can further increase the predictive capability(1.2\%$\uparrow$ KD). Individual components contribute to accuracy, with the MLP alone providing substantial KD and SP scores. The collective employment of all parts in ParZC is essential for the highest prediction accuracy.

\noindent\textbf{Effectiveness of DiffKendall.} We present an ablation study evaluating the impact of Mean Squared Error (MSE) Loss, Rank Loss, and Differentiable Kendall's Tau (DiffKendall) on NB101, as shown in Table \ref{tab:ablation_diffkd}. The results show that employing DiffKendall as the single loss function achieves the best rank correlation with 66.83\% KD and 85.97\% SP. When combined with MSE or Rank loss, the rank correlation deteriorates.
Integrating all three loss functions fails to produce optimal results, highlighting the importance of prioritizing relative scoring in ZC proxies.

\noindent\textbf{Visualization of Rank Correlation.} In Figure~\ref{fig:vis_corr}, we visualize the rank correlation of ParZC on NB201. The left figure displays the correlation across the entire search space, exhibiting a remarkable Kendall Tau value of 78.24\%. We investigate the top-tier architecture within the search space in the right figure and provide a visualization of its correlation. Notably, we mark the top architectures with a star symbol, which validates our ParZC's effectiveness in identifying architectures with superior performance.

\section{Conclusion} 
We present a Parametric Zero-Cost Proxies (ParZC) framework designed to address the critical issue of indiscriminate treatment of node-wise ZC statistics. Specifically, we propose Mixer Architecture with Bayesian Network (MABN) to explore and quantify the inherent uncertainties in the node-wise ZC statistics. To enhance the ranking capabilities of ParZC, we further introduce DiffKendall to handle the discrepancy in ranking architectures. Extensive experiments on various NAS benchmarks and Vision Transformer demonstrate that our ParZC can outperform ZC proxies and predictor-based NAS methods. We aspire for our work to catalyze the design and development of ZC proxies, thereby fostering innovation and progress within the research community.

\clearpage
% \title{Supplementary Materials for \emph{ParZC: Parametric Zero-Cost Proxies for Efficient NAS}}
\setcounter{page}{1}
\maketitlesupplementary
\appendix

\section{Experimental Details}
\label{sec:supp-details}

In this section, we provided further details of experimental settings, including the GBDT experiment described in Section~\ref{sec:intro}, and details of ParZC parameters in NAS-Bench-101 and NAS-Bench-201. 

\subsection{Details of Node-wise ZC statistics}
\label{sec:node-wise-zc}
In this section, we detail the methodology for collecting Node-wise Zero-Cost (ZC) statistics from various NAS benchmarks, including NAS-Bench-101, 201 and NDS. Architectures within these benchmarks are formulated as Directed Acyclic Graphs (DAGs), wherein each node corresponds to a specific operation. Our focus in ParZC predominantly lies on parameter-based nodes, such as Convolutional and Linear layers. We opt to exclude skip connections from our analysis due to practical constraints associated with the inability to collect gradients from these parameter-free nodes. Depth-First Search (DFS) is employed as the primary mechanism for gathering detailed statistical information on the architectures, as delineated in Algorithm~\ref{algo:dfs}.

\begin{algorithm}
\caption{Collection of Node-wise Zero-Cost (ZC) Statistics via Depth-First Search (DFS)}
\label{algo:dfs}
\begin{algorithmic}[1]

\Procedure{CollectZCStatistics}{$DAG$}
    \State $stats \gets [ ]$ 
    % \Comment{Initialize an empty list for statistics}
    \For{each node $n$ in $DAG$}
        \If{$n$ is a parameter-based node}
            \State $dfsStats \gets$ \Call{DFS}{$n$}
            \State $stats$.append($dfsStats$)
            % \Comment{Append DFS statistics to the list}
        \EndIf
    \EndFor
    \State \textbf{return} $stats$
\EndProcedure

\Procedure{DFS}{$node$}
    \State $nodeStats \gets [ ]$ 
    % \Comment{Initialize an empty list for node statistics}
    \State Mark $node$ as visited
    \For{each child $c$ of $node$}
        \If{not visited($c$) and $c$ is parameter-based}
            \State $childStats \gets$ \Call{DFS}{$c$}
            \State Merge $childStats$ into $nodeStats$
        \EndIf
    \EndFor
    \State Compute and append statistics for $node$ to $nodeStats$
    \State \textbf{return} $nodeStats$
\EndProcedure

\end{algorithmic}
\end{algorithm}

For zero-cost proxies, we adopt Fisher~\cite{Turner2019BlockSwapFB_fisher}, GradNorm~\cite{abdelfattah2021zerocost}, GraSP~\cite{Wang2020PickingWT_GraSP}, L2Norm~\cite{abdelfattah2021zerocost}, Plain, SNIP~\cite{Lee2018SNIPSN}, Synflow~\cite{tanaka2020pruning_synflow} for NAS-Bench-101, 201 and NDS. The implementation of these proxies are adopted from ZCNAS~\cite{abdelfattah2021zerocost}. The batch size of calculating these zero-cost proxies is set to 16. Due to the magnitude difference of different ZC statistics, we normalize them to 0-1 range with min-max scaling with max=1 and min=0. We also record the corresponding ground truth performance on test set of the 200-th epoch based on the official NAS benchmarks. For NAS-Bench-201, we use the up-to-date NAS-Bench-201-v1\_1-096897.pth as the benchmark file.

\subsection{Details of GBDT}
\begin{figure}
    \centering
    \includegraphics[width=1\linewidth]{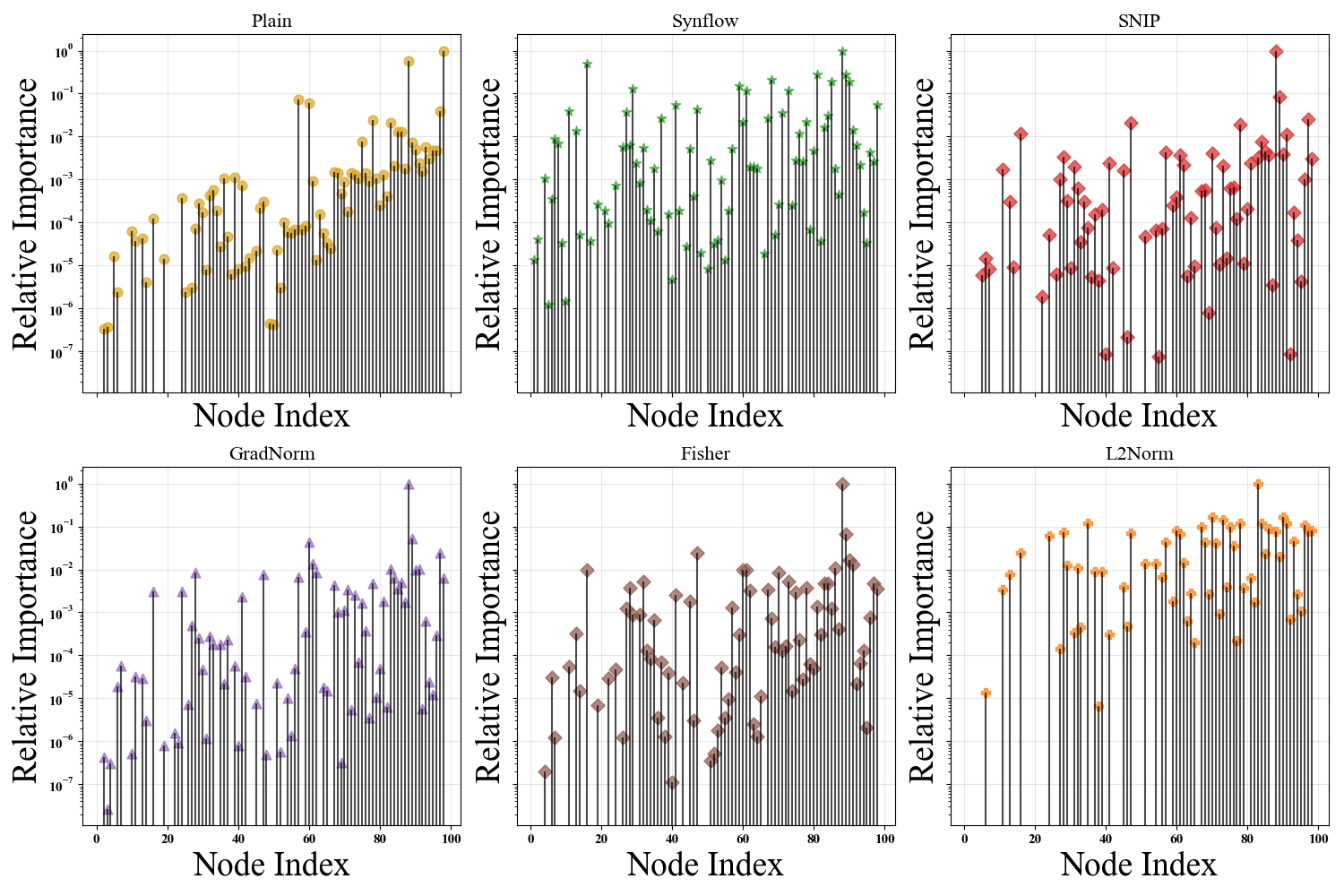}
    \caption{Visualization of node-wise importance of different ZC proxies as input.}
    \label{fig:all_gbdt_zc}
\end{figure}

The Gradient Boosting Decision Tree (GBDT), a machine learning algorithm renowned for its efficiency and interpretability, is employed in our study for the in-depth analysis of node-wise Zero-Cost (ZC) proxies. The choice of GBDT is motivated by its superior interpretability, making it an ideal tool for assessing the importance and contribution of various nodes within a network architecture. Utilizing impurity-based measures, GBDT facilitates the visualization of the differential contributions of network layers. As illustrated in Figure~\ref{fig:all_gbdt_zc}, our analysis extends beyond the insights gleaned from Figure~\ref{fig:insight} by including three additional ZC proxies, namely Plain, Synflow, SNIP, GradNorm, Fisher, and L2Norm.

In line with the data collection procedures outlined in Section~\ref{sec:node-wise-zc}, we partition the dataset into an 80\% training set and a 20\% test set. This division ensures a robust training process while allowing for an accurate evaluation of the model's performance. The configuration of the GBDT model in our study is carefully selected to optimize its effectiveness. It includes 500 estimators, offering a comprehensive and nuanced understanding of the data. The learning rate is set to 0.05, balancing the speed of learning with the risk of overfitting. The maximum depth of each tree in the GBDT is capped at 3, a choice that aids in preventing overfitting while maintaining model simplicity for easier interpretation. Finally, the random state is set to 42, ensuring consistency and reproducibility in our results.

Through this meticulous application of GBDT, we aim to present a detailed and insightful analysis of the node-wise ZC proxies, contributing significantly to the understanding of imbalanced contribution of node-wise ZC statistics.

\subsection{Details Setting on NAS Benchmarks}
% Here, we clearify that due to the maturity of tools chains in NAS-Bench-101 and NAS-Bench-201 as provided in PINAT~\cite{PINATAAAI23}, we treat the architectural information as the complementary information to node-wise ZC statistics. To be specific, we adopt the pipeline of PINAT, which takes the adjancy matrix to represent the architecture information and operations vector as the operation information. Therefore, the embedding of PINAT has architectural and operation informations, but lack more detailed information as implied in node-wise ZC statistics. We combine the embedding of PINAT and our proposed ParZC together to predict the rank of different architectures. But for NDS, the search space of ENAS, NASNet, etc, is more practical than that of NAS-Bench-101 and 201, thus having a more complicated architecture, which is hard to represented by adjacency matrix. Therefore, in this work, we directly use the node-wise zero-cost statistics as input and not employ any architectual information as additional input. But results can demonstrate that our proposed zero-cost proxies can still outperform that of exisiting zero-cost proxies as detailed in Table~\ref{tab:nb101_201_nds}.

In this work, we clarify that due to the advanced tool chains in NAS-Bench-101 and NAS-Bench-201, as detailed in PINAT~\cite{PINATAAAI23}, we consider the architectural information as supplemental to the node-wise Zero-Cost (ZC) statistics. Specifically, we utilize the PINAT pipeline, which employs an adjacency matrix to represent architectural information and an operations vector for operation details. Consequently, while the PINAT embedding encompasses architectural and operational data, it lacks the granular insights provided by node-wise ZC statistics. However, in the case of NDS, encompassing search spaces like ENAS and NASNet, the architectures are more complex than those in NAS-Bench-101 and 201, and thus challenging to represent solely with an adjacency matrix. Therefore, in our approach, we exclusively rely on node-wise zero-cost statistics as our input, without incorporating any additional architectural information. Our results, as detailed in Table~\ref{tab:nb101_201_nds}, demonstrate that our proposed zero-cost proxies outperform existing zero-cost proxies, underscoring their efficacy in more practical and complex scenarios.

\begin{figure*}[t]
    \centering
    \begin{subfigure}[b]{0.48\textwidth}
        \centering
        \includegraphics[width=\linewidth]{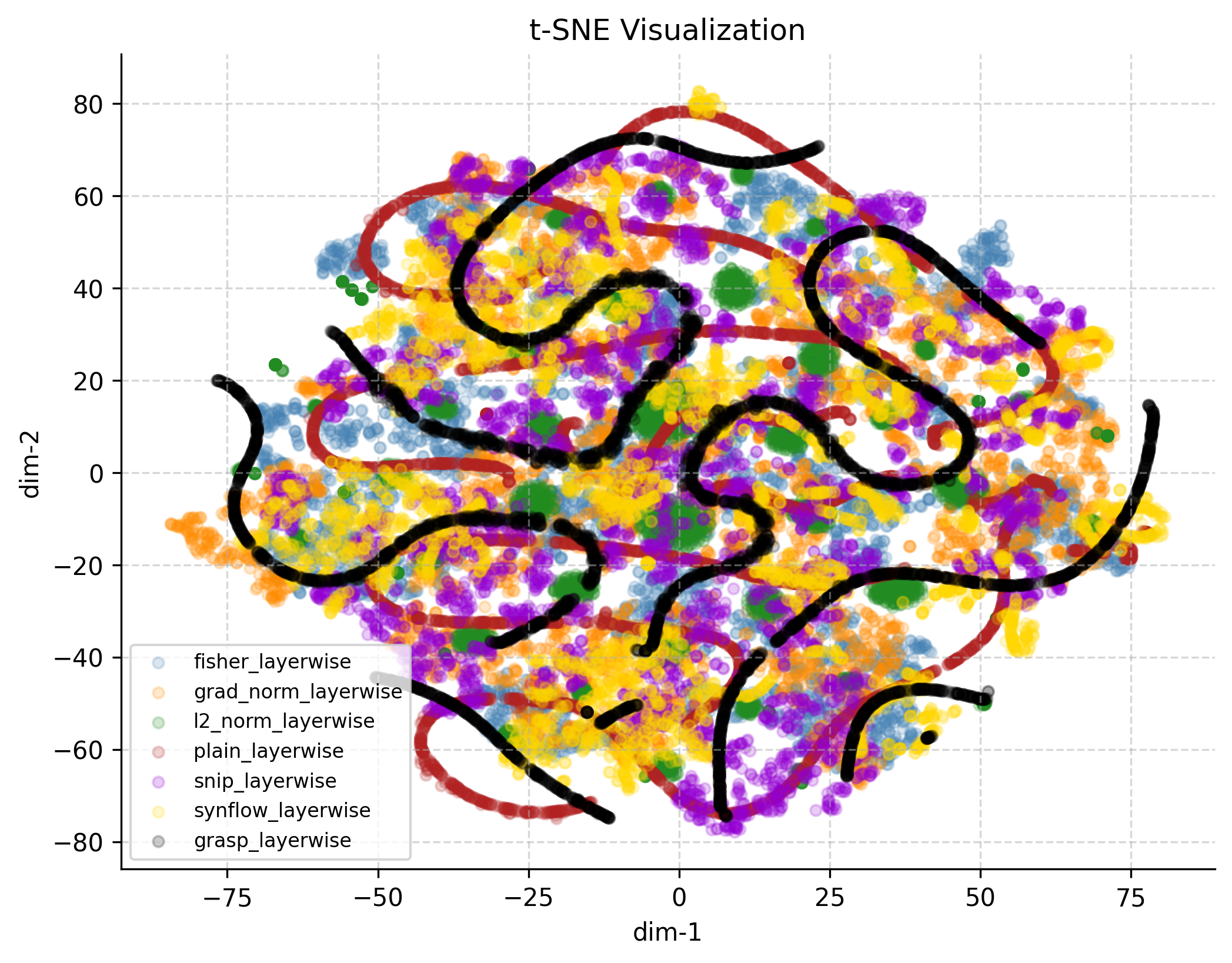}
        \caption{T-SNE visualization of different ZC on NAS-Bench-101}
        \label{fig:tsne-nb101}
    \end{subfigure}
    \hfill
    \begin{subfigure}[b]{0.48\textwidth}
        \centering
        \includegraphics[width=\linewidth]{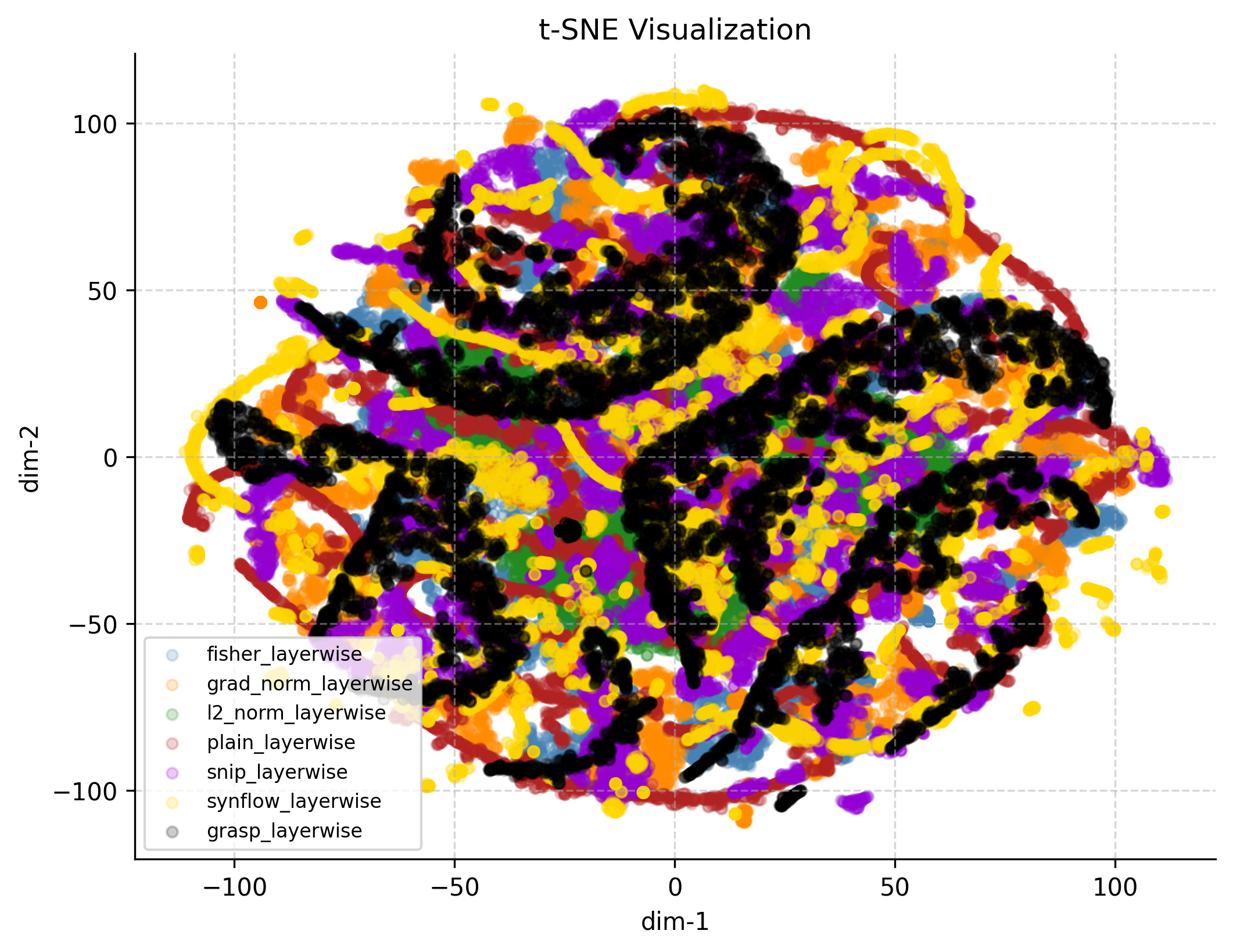}
        \caption{T-SNE visualization of different ZC on NAS-Bench-201}
        \label{fig:tsne-nb201}
    \end{subfigure}
    \caption{Parallel T-SNE visualizations of different ZC on NAS-Bench-101 and NAS-Bench-201}
    \label{fig:tsne-parallel}
\end{figure*}

\subsection{Encoding Architectural Information}
To further make up for the inability to perceive architectural shortcomings, we proposed to inject the architecture information using a unified rule with an adjacency matrix $\mathcal{A} \in \mathbb{R}^{N \times N}$ and operations encoding $V \in \mathbb{R}^{N \times 1}$, where each architecture is regarded as a Directed Acyclic Graph (DAG) structure. Following the standard encoding proposed by NP\cite{wen2020neural}, we adopted GCN to encode the architecture information. The encoding process begins by representing each architecture as a graph in which nodes correspond to operations or nodes, and edges represent the flow of data between these components. The adjacency matrix $\mathcal{A}$ captures the connectivity between nodes, while the operations encoding $V$ represents the specific operation at each node. This graph-based representation is crucial as it captures both the structural and functional aspects of the architecture. The Graph Convolutional Network (GCN) is then employed to encode this graph representation into a feature space. The GCN, through its convolutional layers, aggregates information from the neighbors of each node, effectively capturing the local and global structural properties of the architecture. The update rule for a layer in GCN can be formulated as:
\begin{equation}
    H^{(l+1)} = \sigma\left(\hat{D}^{-\frac{1}{2}}\hat{A}\hat{D}^{-\frac{1}{2}}H^{(l)}W^{(l)}\right),
\end{equation}
where $\hat{A} = \mathcal{A} + I_N$ is the adjacency matrix with added self-connections, $\hat{D}$ is the degree matrix of $\hat{A}$, $H^{(l)}$ is the activation in the $l$-th node, $W^{(l)}$ is the weight matrix for the $l$-th node, and $\sigma(\cdot)$ is the non-linear activation function.  Then we follow the Transformer architecture with Permutation invariance module proposed by PINAT~\cite{PINATAAAI23} to further fuse the architectural information. We denote the weight of Transformer as $W^{(p)}$, then we get the embedding $H^{(p)}=W^{(p)}(\hat{A}, V)$. For the ParZC denoted as $W^{(z)}$, we take the node-wise ZC statistics $\mathcal{Z}$ as input and get the embedding $H^{z}=W^{(z)}(\mathcal{Z})$. Then we further fuse them by adding them element-wise $H=H^{(z)}+H^{(p)}$. After that, we employ a regressor $\mathcal{R}$ to generate the final ranking score $y=\mathcal{R}(H)$, where the regressor is two fully connected linear layers. 
 
This  encoding allows us to represent the architecture in a form that is amenable to analysis by downstream learning algorithms. It enables the model to not only understand the individual components of the architecture but also the complex inter-dependencies between these components. Consequently, this approach facilitates a more nuanced and informed search process in the neural architecture search paradigm, potentially leading to the discovery of more efficient and effective neural network architectures.

For the subsequent NAS benchmarks, we have employed Optuna to conduct an exhaustive search for the optimal hyperparameters of the models. Below, we present the search space specifications for the ParZC model: The search space for the ParZC model encompasses a diverse array of hyperparameters that collectively dictate its architectural configuration and training attributes. These hyperparameters encompass the following: the number of layers (\texttt{n\_layers}), which can assume values within the range of 2 to 5; the number of attention heads $(\texttt{n\_head})$, which spans from 3 to 8; the hidden layer size within the Pine component $(\texttt{pine\_hidden})$, with values ranging from 8 to 128; the dimensionality of word representations $(\texttt{d\_word\_model})$, which varies between 256 and 1024; the dimensionality of keys and values in the attention mechanism $(\texttt{d\_k\_v})$, which lies in the interval of 32 to 128; the dimensionality of inner layers $(\texttt{d\_inner})$, ranging from 256 to 1024; the number of training epochs $(\texttt{epoch})$, spanning from 20 to 300; and the dropout rate $(\texttt{dropout})$, which assumes values between 0.01 and 0.5. Collectively, these hyperparameters govern the ParZC model's architectural depth, the behavior of its attention mechanism, the characteristics of its hidden layers, the encoding of word representations, and the specifics of its training regimen. This expansive search space is meticulously designed to facilitate optimization tailored to the unique demands of specific tasks and datasets.

\textbf{Details for NAS-Bench-101}
% For NAS-Bench-101, we employ a similar methodology to gather node-wise statistics. Specifically, we utilize Synflow, Snip, GradNorm, and Fisher as proxies to generate statistics for different architectural configurations within the neural network. To account for variations in the number of parameter-based operations across different architectures, we pad each generated list with zeros to ensure a consistent maximum length. By concatenating the lists produced by these different Zero-Cost (ZC) proxies, we obtain a total dimension of 249. We then proceed to sample 1000 architectures from NAS-Bench-101, considering architectures from ENAS, DARTS, and NASNet. This dataset is subsequently divided into a 60\% training set and a 40\% validation set. For the training process, we employ ParZC with a segment length of 752 and a segment size of 16. Information fusion is carried out using five segment mixers, and a dropout rate of 0.18 is applied. Additionally, we set the expansion factor to 4 and the expansion token factor to 0.5 to configure the model appropriately for the task.

For the NAS-Bench-101 experiments, we followed a similar methodology to collect node-wise statistics. Specifically, we utilized Synflow, Snip, GradNorm, and Fisher as proxies to generate statistics for different architectural configurations within the neural network. To accommodate variations in the number of parameter-based operations across architectures, we padded each generated list with zeros to ensure a consistent maximum length. By concatenating the lists produced by these Zero-Cost (ZC) proxies, we obtained a unified dataset with a total dimension of 249.

Subsequently, we sampled 1000 architectures from the NAS-Bench-101 dataset, specifically considering architectures from ENAS, DARTS, and NASNet. The sampled dataset was then divided into a 60\% training set and a 40\% validation set for subsequent analysis.
During the training process, we utilized the ParZC model with a segment length of 752 and a segment size of 16. Information fusion was performed using five segment mixers, and a dropout rate of 0.18 was applied for regularization.

Additionally, we configured the model with an expansion factor of 4 and an expansion token factor of 0.5 to adapt it appropriately for the task at hand. The ParZC model, denoted as net, consisted of four layers, six attention heads, a Pine hidden size of 8, and a linear hidden size of 708. It also had a source vocabulary size of 5, with a word vector dimensionality of 708. The dimensions for keys (d\_k) and values (d\_v) were set to 100, while the overall model dimension (d\_model) was set to 708. The inner dimension (d\_inner) was configured as 530.

\textbf{Details for NAS-Bench-201}
% For NAS-Bench-201, we adopt a similar methodology to collect architectural statistics. Specifically, we utilize Synflow, Snip, GradNorm, and Fisher as proxies to generate node-wise statistics for various architectural configurations within the neural network. Given the variability in the number of parameter-based operations across different architectures, we pad each generated list with zeros to ensure uniformity in list length. By concatenating the lists produced by these different Zero-Cost (ZC) proxies, we create a unified dataset with a total dimension of 2832. Subsequently, we proceed to sample 1000 architectures from NAS-Bench-201, spanning a range of architectural designs. This dataset is then split into a 60\% training set and a 40\% validation set for subsequent analysis. During training, we employ the ParZC model with a segment length of 752 and a segment size of 16. Information fusion is facilitated using five segment mixers, and a dropout rate of 0.18 is applied for regularization. Additionally, we configure the model with an expansion factor of 4 and an expansion token factor of 0.5 to tailor it to the specific requirements of the task at hand.

For the NAS-Bench-201 experiments, we followed a similar methodology to gather architectural statistics. Specifically, we employed Synflow, Snip, GradNorm, and Fisher as proxies to generate node-wise statistics for different architectural configurations within the neural network. To account for the variability in parameter-based operations across architectures, we padded each generated list with zeros to ensure consistent list lengths. By concatenating the lists obtained from these Zero-Cost (ZC) proxies, we constructed a unified dataset with a total dimension of 294.

Subsequently, we randomly sampled 1,000 architectures from the NAS-Bench-201 dataset, encompassing a diverse range of architectural designs. The sampled dataset was then divided into a 60\% training set and a 40\% validation set for subsequent analysis. 
During the training process, we utilized the ParZC model with a segment length of 752 and a segment size of 16. Information fusion was facilitated using five segment mixers. To prevent overfitting, we applied a dropout rate of 0.18 for regularization.

Furthermore, we configured the model with an expansion factor of 4 and an expansion token factor of 0.5, tailoring it to the specific requirements of the task at hand. The architecture of the ParZC model, denoted as net, included four layers, six attention heads, a Pine hidden size of 76, and a linear hidden size of 765. It also had a source vocabulary size of 5, with word vector dimensionality of 765. The dimensions for keys (d\_k) and values (d\_v) were set to 100, while the model's overall dimension (d\_model) was set to 765. The inner dimension (d\_inner) was configured as 338.

\begin{table}[b]
\centering
\caption{Ranking correlation of ParZC with different modules under different seeds.}
\resizebox{\linewidth}{!}{
\begin{tabular}{cccccccccc}
\toprule
   &       &    &     & \multicolumn{2}{c}{Run1} & \multicolumn{2}{c}{Run2} & \multicolumn{2}{c}{Run3} \\ \midrule
NP & Mixer & BN & MLP & KD          & SP         & KD          & SP         & KD          & SP         \\ \midrule
$\checkmark$  & -     & -  & -   & 33.66       & 44.35      & 35.19       & 50.50      & 34.03       & 41.98      \\
-  & $\checkmark$     & -  & -   & 64.74       & 79.32      & 60.38       & 79.36      & 61.76       & 81.89      \\
-  & -     & -  & $\checkmark$   & 57.06       & 75.85      & 50.51       & 68.26      & 56.34       & 76.39      \\
-  & -     & $\checkmark$  & -   & 50.57       & 68.85      & 47.09       & 65.57      & 52.70       & 72.85      \\
$\checkmark$  & $\checkmark$     & -  & -   & 57.89       & 76.35      & 59.80       & 78.59      & 61.69       & 81.59      \\
$\checkmark$  & -     & $\checkmark$  & -   & 56.02       & 75.03      & 53.35       & 72.51      & 55.06       & 75.19      \\
$\checkmark$  & $\checkmark$     & $\checkmark$  & -   & 69.98       & 87.90      & 67.24       & 86.06      & 69.44       & 87.56      \\ \bottomrule
\end{tabular}}\label{tab:seeds}
\end{table}

\begin{table*}[t]
  \centering
    \caption{Search results on NAS-Bench-201. The standard deviation is in the subscript.}
 \resizebox{170mm}{!}{
  \begin{tabular}{c|c|cc|cc|cc}
    \toprule
	\multirow{2}{*}{\textbf{Method}} &\multirow{2}{*}{\textbf{search seconds}}
	& \multicolumn{2}{c|}{\textbf{CIFAR-10 (\%)}} & \multicolumn{2}{c|}{\textbf{CIFAR-100 (\%)}} & \multicolumn{2}{c}{\textbf{ImageNet-16-120 (\%)}}\\
    \cline{3-8}
    & & valid & test & valid & test & valid & test\\
	\hline 
	\midrule
   RSPS~\cite{dong2019NASBench201} & 7587.12 &  84.16$_{(1.69)}$ & 87.66$_{(1.69)}$ & 59.00$_{(4.60)}$ & 58.33$_{(4.34)}$ & 31.56$_{(3.28)}$ & 31.14$_{(3.88)}$ \\
   %DARTS-V1~\cite{liu2018darts} & 10889.87 &  39.77$_{(0.00)}$ & 54.30$_{(0.00)}$ & 15.03$_{(0.00)}$ & 15.61$_{(0.00)}$  & 16.43$_{(0.00)}$ & 16.32$_{(0.00)}$ \\
   DARTS-V2~\cite{darts} & 29901.67 & 39.77$_{(0.00)}$ & 54.30$_{(0.00)}$ & 15.03$_{(0.00)}$ & 15.61$_{(0.00)}$ & 16.43$_{(0.00)}$ & 16.32$_{(0.00)}$ \\
   GDAS~\cite{dong2019search} &  28925.91 & 90.00$_{(0.21)}$ & 93.51$_{(0.13)}$ & 71.15$_{(0.27)}$ & 70.61$_{(0.26)}$ & 41.70$_{(1.26)}$ & 41.84$_{(0.90)}$ \\
   SETN~\cite{SETN} & 31009.81 &  82.25$_{(5.17)}$ & 86.19$_{(4.63)}$ & 56.86$_{(7.59)}$ & 56.87$_{(7.77)}$ & 32.54$_{(3.63)}$ & 31.90$_{(4.07)}$ \\
   ENAS-V2~\cite{pham2018efficient} & 13314.51 & 39.77$_{(0.00)}$ & 54.30$_{(0.00)}$ & 15.03$_{(0.00)}$ & 15.61$_{(0.00)}$ & 16.43$_{(0.00)}$ & 16.32$_{(0.00)}$ \\
   \hline
   \midrule
   Random Sample & 0.01 & 90.03$_{(0.36)}$ & 93.70$_{(0.36)}$ & 70.93$_{(1.09)}$ & 71.04$_{(1.07)}$ & 44.45$_{(1.10)}$ & 44.57$_{(1.25)}$ \\
   NPENAS~\cite{wei2022npenas} & - & 91.08$_{(0.11)}$ & 91.52$_{(0.16)}$ & - & - & - & - \\
   REA~\cite{real2018regularized} & 0.02 & 91.19$_{(0.31)}$ & 93.92$_{(0.30)}$ & 71.81$_{(1.12)}$ & 71.84$_{(0.99)}$ & 45.15$_{(0.89)}$ & 45.54$_{(1.03)}$ \\
   NASBOT~\cite{kandasamy2018neural} %& 90 query 
   & -    &  - & 93.64$_{(0.23)}$ & - & 71.38$_{(0.82)}$ & - & 45.88$_{(0.37)}$ \\
   REINFORCE~\cite{williams1992Simple} & 0.12 &  91.09$_{(0.37)}$ & 93.85$_{(0.37)}$ & 71.61$_{(1.12)}$ & 71.71$_{(1.09)}$ & 45.05$_{(1.02)}$ & 45.24$_{(1.18)}$ \\
   BOHB~\cite{falkner2018bohb} & 3.59 &  90.82$_{(0.53)}$ & 93.61$_{(0.52)}$ & 70.74$_{(1.29)}$ & 70.85$_{(1.28)}$ & 44.26$_{(1.36)}$ & 44.42$_{(1.49)}$ \\
   ReNAS~\cite{xu2021renas} %& 90 query 
   & 86.31 & 90.90$_{(0.31)}$ & 93.99$_{(0.25)}$ & 71.96$_{(0.99)}$ & 72.12$_{(0.79)}$ & 45.85$_{(0.47)}$ & 45.97$_{(0.49)}$ \\
   %\revise{ProxyBO~\cite{shen2021proxybo}} & - & \revise{91.44$_{(0.10)}$} & - & - & \revise{73.48$_{(0.17)}$} & - & -\\
   %\revise{AceNAS~\cite{zhang2021acenas}} & - & - & - & - & \revise{73.47} & - & \revise{46.34}\\
   	\midrule
   ParZC(Ours) & 68.95 & \textbf{91.55}$_{(0.02)}$ %\textbf{91.57}$_{(0.03)}$ 
   & \textbf{94.36}$_{(0.01)}$ &\textbf{73.49}$_{(0.02)}$ & \textbf{73.51}$_{(0.00)}$  &\textbf{46.37}$_{(0.04)}$  & \textbf{46.34}$_{(0.01)}$  \\
   \hline
   \midrule
   Optimal & - & 91.61 & 94.37 & 73.49 & 73.51 & 46.73 & 47.31 \\
   \cline{1-1}\cline{2-8}
   \midrule
   ResNet  & - & 90.83 & 93.97 & 70.42 & 70.86 & 44.53 & 43.63 \\ 
   \bottomrule
\end{tabular}
}
 \label{tab:nb201_search_result}
\end{table*}

\textbf{Details for NDS} We utilize Synflow, Snip, GradNorm, and Fisher as zero-cost (ZC) proxies to generate node-wise statistics for NDS. Since the number of parameter-based operations in different architectures within a neural network can vary, we pad the lists generated by different ZCs with zeros to ensure they have the same maximum length.
By concatenating the lists generated by different zero-cost proxies (ZCs), we obtain a total dimension of 2832 for the Amoeba search space. The input dimension varies across different search spaces as follows: 2,832 for Amoeba, 2,000 for DARTS, 2,752 for ENAS, 2,520 for NASNet, and 2,912 for PNAS.

We randomly sample 1,000 architectures from the NDS on ENAS, DARTS, and NASNet, and split them into a 60\% training set and a 40\% validation set. For our ParZC approach, we employ a segment length of 752 with a segment size of 16, and we use five segment mixers to fuse the information. The dropout rate is set to 0.18, while the expansion factor and expansion token factor are set to 4 and 0.5, respectively. The `Time' column in the resulting table indicates the evaluation time (in seconds) for each bit-width configuration.

\begin{table}
\centering
\caption{Hyperparameter (HP) Search Space of Optuna.}
\begin{tabular}{l|c}
  \toprule
  \textbf{HP} & \textbf{Value} \\
  \midrule
  Patch Size & 16 \\
  \hline
  Max Seq Length & 4096 \\
  \hline
  Dimension & $[256, 4096]$ \\
  \hline
  Depth & $[2, 8]$ \\
  \hline
  Dropout & $[0.1, 0.5]$ \\
  \hline
  Batch Size & $[16, 128]$ \\
  \hline
  Epochs & $[50, 300]$ \\
  \hline
  Learning Rate & $[1e-4, 1e-2]$\\
  \bottomrule
\end{tabular}
\end{table}

\section{Extended Experiments on MQ-Bench-101}
To evaluate the generalization ability of ParZC, we conducted extended experiments on mixed-precision quantization (MQ) using the MQ-Bench-101 benchmark from EMQ~\cite{dong2023emq}. 
MQ-Bench-101 is specifically designed to assess the performance of different bit configurations in post-training quantization on ResNet-18, considering various bit-widths for weights and activations. This benchmark enables the comparison of different MQ proxies in terms of their rank consistency and predictive ability.
Table~\ref{tab:mq-bench} presents the results of the rank correlation analysis (\%) for training-free proxies on MQ-Bench-101. The $\mathit{Spearman@topk} (\rho_{s@k})$ metric is used to measure the correlation of the top performing bit configurations on the benchmark. The table includes various methods such as BParams, HAWQ, HAWQ-V2, OMPQ, QE, SNIP, Synflow, EMQ, and our proposed ParZC. The results show the rank correlation values for different top-k percentages ($20\%$, $50\%$, and $100\%$). Additionally, the evaluation time (in seconds) for each method is provided in the `Time(s)' column. Notably, our proposed ParZC demonstrates comparable performance to the state-of-the-art EMQ method~\cite{dong2023emq} in terms of rank correlation on MQ-Bench-101.

\section{Detailed Performance on NAS-Bench-201}

In this section, we provide additional details on the performance of our ParZC method on both the test set and validation set of NAS-Bench-201. While the main paper presents the results on the test set, we aim to further verify the effectiveness of our ParZC method by providing a more comprehensive analysis.

As shown in Table~\ref{tab:sota-nasbench201}, ParZC consistently outperforms both training-based and training-free baselines by selecting architectures with superior performance. The search process of ParZC only requires 68 GPU seconds since it estimates the performance of architectures in batches. This is significantly shorter than even zero-shot NAS methods like GradSign \cite{gradsign}. Furthermore, our ParZC model achieves state-of-the-art accuracy on NAS-Bench-201 with minimal variance, demonstrating its efficiency and effectiveness in finding high-performing architectures.

\section{Stable Analysis}

In this section, we present the results of our ablation study in Table~\ref{tab:seeds}, which involved multiple runs with different seeds to ensure the stability and robustness of our findings. By conducting these experiments with varying seeds, we aimed to investigate the consistency and reliability of our results. The use of different seeds allows us to account for the potential influence of randomness in the experimental process.
Through our ablation study, we carefully examined the impact of specific variables or components by systematically removing or modifying them in each run. By comparing the results across multiple runs, we can assess the stability of our findings and determine the extent to which our conclusions hold under different conditions. 
The use of diverse seeds in our ablation study ensures that our analysis is not biased by a particular seeds. Instead, it provides a more comprehensive understanding of the behavior and performance of our experimental setup.

\section{Diversity of Different Zero-cost Proxies}

We analyze the distribution of various ZC proxies through a visualization technique, as depicted in Figure~\ref{fig:tsne-nb101} and~\ref{fig:tsne-nb201}, using a dataset of 5,000 architectures from NAS-Bench-101 and 15,625 architectures from NAS-Bench-201. This visual representation provides valuable insights into the patterns and characteristics of the ZC proxies employed in our study. In the t-SNE visualization, we observe that different ZC proxies exhibit distinct patterns. For example, the ZC proxies associated with the "plain" and "grasp" architectures form a line-like structure. This indicates that these architectures have similar characteristics or share common design principles. The linear arrangement suggests a gradual progression or transition between these architectures, with slight variations in their features or performance.

On the other hand, the remaining ZC proxies, such as those corresponding to "curve," "spiral," or other architectural variations, are distributed more uniformly across the search space. This distribution implies a higher degree of diversity among these architectures, with each ZC proxy representing a unique design or approach. Unlike the linear arrangement observed in "plain" and "grasp," the absence of a clear pattern among these ZC proxies suggests a wider exploration of architectural possibilities.

The diversity observed in ZC proxies through t-SNE visualization showcases the remarkable versatility and richness of zero-shot NAS. This visualization not only demonstrates the ability of ZC proxies to encapsulate a broad spectrum of architectural characteristics and design choices but also highlights their potential for facilitating efficient exploration of the architectural space without relying on computationally expensive training cycles. This crucial observation serves as the foundation and provides an intuitive understanding for the validity of our proposed ParZC technique. By fully leveraging the exploration of these diverse ZC proxies, we can expect to achieve enhanced performance in neural architecture search. Overall, these profound insights underscore the power and flexibility of zero-cost proxies as an invaluable tool within the field of neural architecture search.

\begin{table}
	\centering
	\small
	\caption{Rank correlation (\%) of training-free proxies on MQ-Bench-101. The $\mathit{Spearman@topk} (\rho_{s@k})$ are adopted to measure the correlation of the top performing bit configurations on MQ-Bench-101.}
	\resizebox{1\linewidth}{!}{
		\begin{tabular}{lrrrr}
			\toprule[1pt]
			Method                              & $\rho_{s@20\%}$    & $\rho_{s@50\%}$    & $\rho_{s@100\%}$   & Time(s) \\
			\midrule
   			BParams                             & 28.67$_{\pm 0.24}$ & 32.41$_{\pm 0.07}$ & 55.08$_{\pm 0.13}$ & 2.59    \\
			HAWQ~\cite{dong2019hawq}            & 23.64$_{\pm 0.13}$ & 36.21$_{\pm 0.09}$ & 60.47$_{\pm 0.07}$ & 53.76   \\
			HAWQ-V2~\cite{dong2019hawqv2}        & 30.19$_{\pm 0.14}$ & 44.12$_{\pm 0.15}$ & 74.75$_{\pm 0.05}$ & 42.17   \\
			OMPQ~\cite{Ma2021OMPQOM}            & 7.88$_{\pm 0.16}$  & 16.38$_{\pm 0.08}$ & 31.07$_{\pm 0.03}$ & 53.76   \\
			QE~\cite{qescore}              & 20.33$_{\pm 0.09}$ & 24.37$_{\pm 0.13}$ & 36.50$_{\pm 0.06}$ & 2.15    \\
			SNIP~\cite{Lee2018SNIPSN}            & 33.63$_{\pm 0.20}$ & 17.23$_{\pm 0.09}$ & 38.48$_{\pm 0.09}$ & 2.50    \\
			Synflow~\cite{tanaka2020pruning_synflow} & 39.92$_{\pm 0.09}$ & 44.10$_{\pm 0.11}$ & 31.57$_{\pm 0.02}$ & 2.23    \\
			EMQ~\cite{dong2023emq}    & \textbf{42.59}$_{\pm 0.09}$ & 57.21$_{\pm 0.05}$ & \textbf{79.21}$_{\pm 0.05}$ & \textbf{1.02}    \\
                ParZC(Ours) & 40.47$_{\pm 0.14}$& \textbf{66.84$_{\pm 0.08}$}  &\textbf{80.05$_{\pm 0.12}$}& 2.3 \\
			\bottomrule[1pt]
		\end{tabular}}
	\label{tab:mq-bench}%
 \vspace{-3em}
\end{table}

\section{More Visualization Results}

\subsection{Visualization of Bayesian Network}
\begin{figure}
    \centering
    \includegraphics[width=1\linewidth]{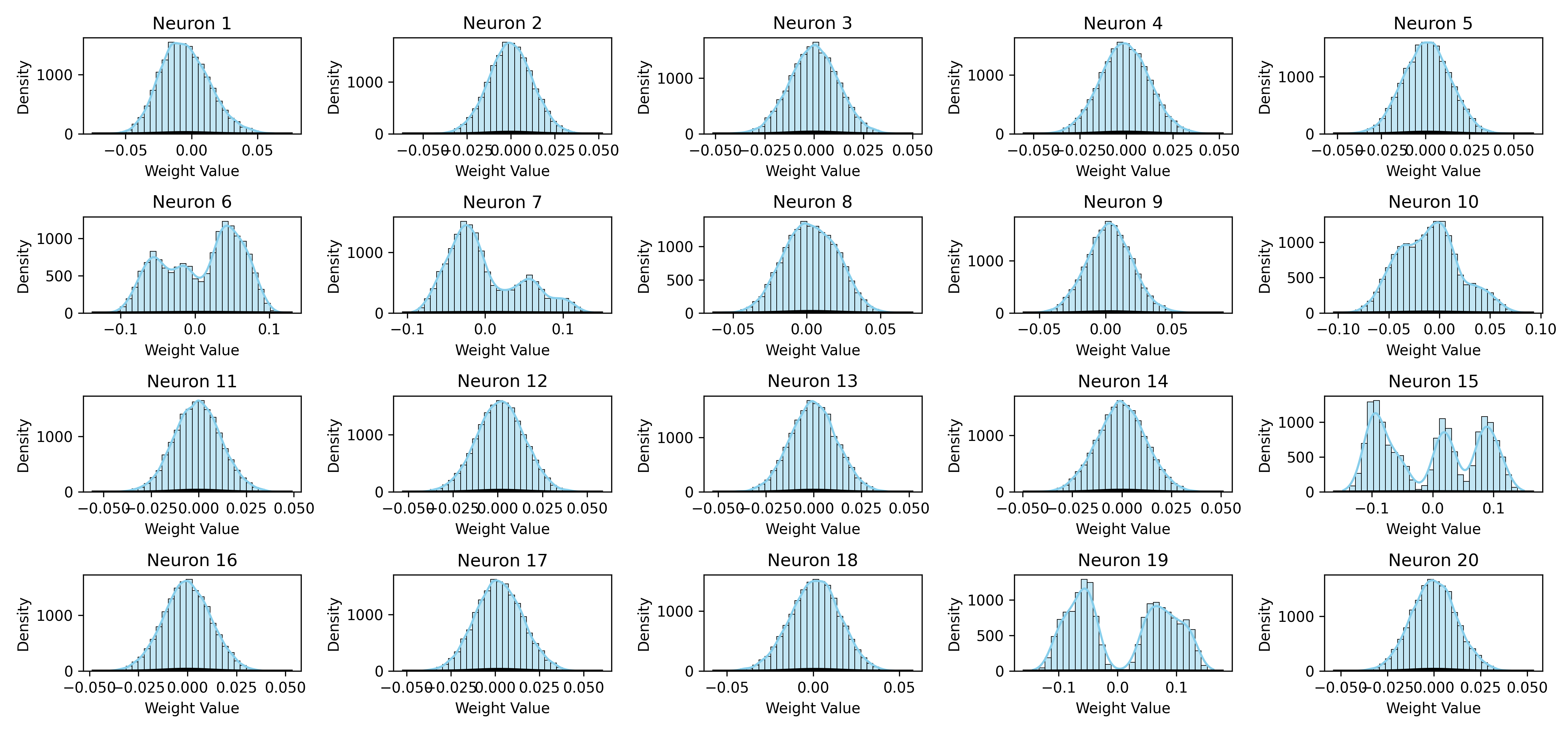}
    \caption{Visualization of Bayesian Network}
    \label{fig:bayesian_network}
\end{figure}

Figure~\ref{fig:bayesian_network} presents a comprehensive visualization of the distributions of weight values across the neurons in our Bayesian Network. Each sub-figure, from Neuron 1 to Neuron 21, contains a histogram that elucidates the frequency distribution of weights, providing an empirical basis to analyze the uncertainty associated with the zero-cost proxies utilized within the network. 

The horizontal axis (x-axis) denotes the weight values, while the vertical axis (y-axis) corresponds to the frequency of these values. This alignment of histograms enables a parallel comparison among the neurons, highlighting the variability and consistency of the learned parameters. Such a visualization is instrumental in deciphering the dispersion and central tendencies within the network, which are pivotal for a nuanced understanding of the uncertainty encapsulated by the ZC proxies.

\subsection{Correlation Visualization of ParZC on NAS Benchmarks}

\begin{figure*}
  \centering
  \begin{subfigure}{0.24\textwidth}
    \includegraphics[width=\linewidth]{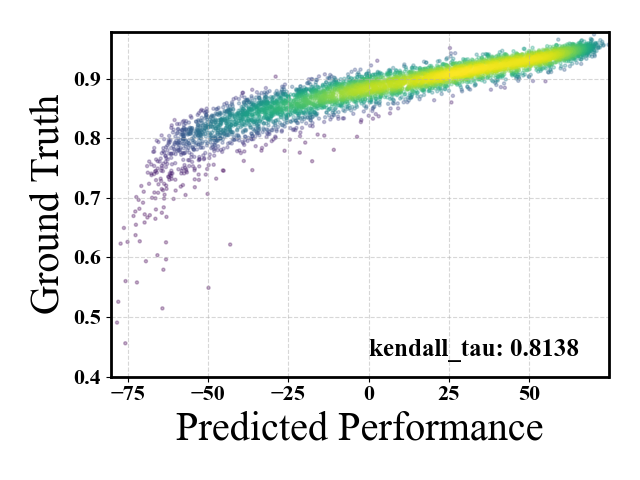}
    \caption{NB101: CIFAR-10}
  \end{subfigure}
  \begin{subfigure}{0.24\textwidth}
    \includegraphics[width=\linewidth]{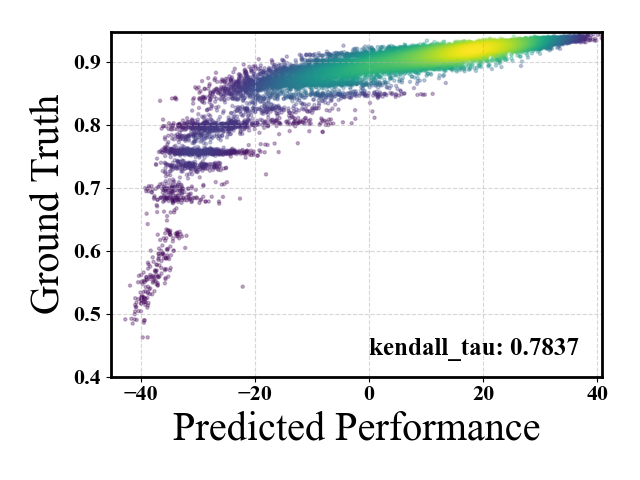}
    \caption{NB201: CIFAR-10}
  \end{subfigure}
  \begin{subfigure}{0.24\textwidth}
    \includegraphics[width=\linewidth]{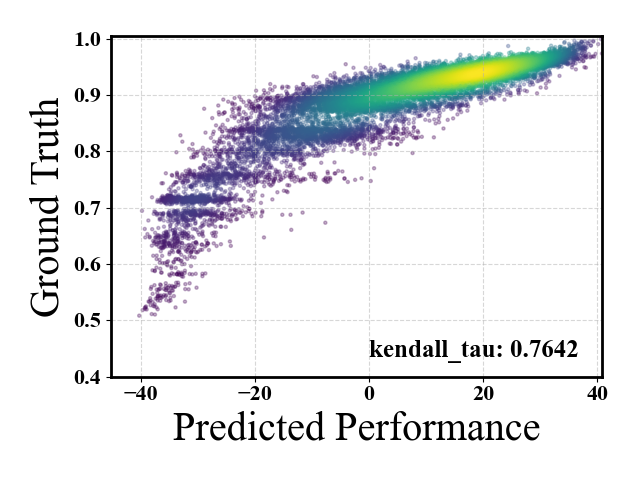}
    \caption{NB201: CIFAR-100}
  \end{subfigure}
  \begin{subfigure}{0.24\textwidth}
    \includegraphics[width=\linewidth]{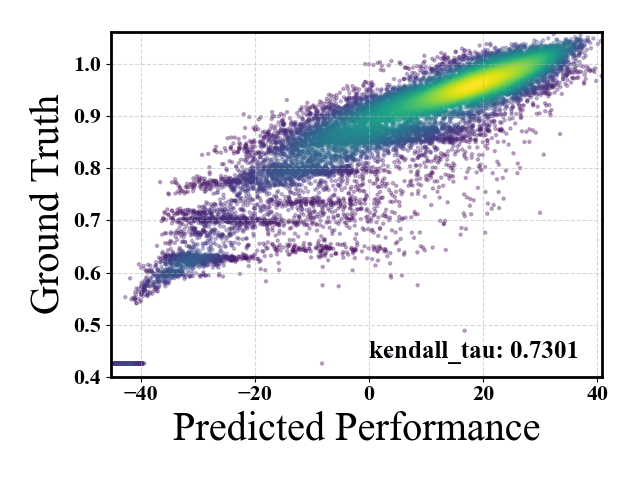}
    \caption{NB201: ImageNet16}
  \end{subfigure}
  \\
  \begin{subfigure}{0.24\textwidth}
    \includegraphics[width=\linewidth]{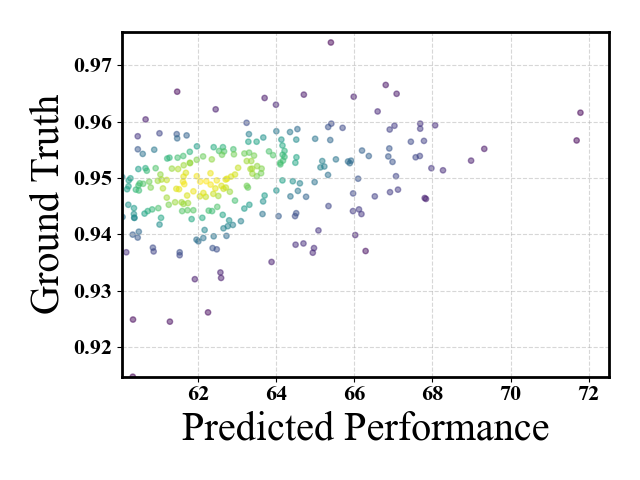}
    \caption{NB101: CIFAR-10 (Top 5\%)}
  \end{subfigure}
  \begin{subfigure}{0.24\textwidth}
    \includegraphics[width=\linewidth]{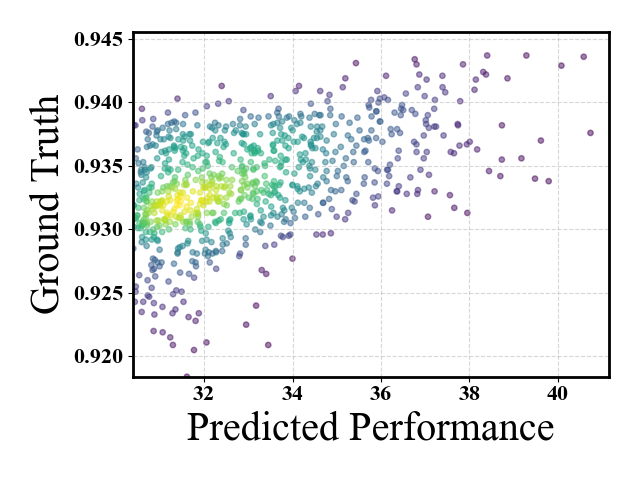}
    \caption{NB201: CIFAR-10 (Top 5\%)}
  \end{subfigure}d
  \begin{subfigure}{0.24\textwidth}
    \includegraphics[width=\linewidth]{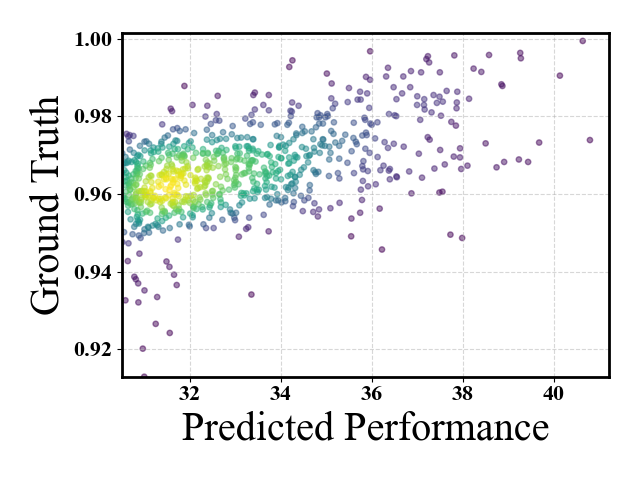}
    \caption{NB201: CIFAR-100 (Top 5\%)}
  \end{subfigure}
  \begin{subfigure}{0.24\textwidth}
    \includegraphics[width=\linewidth]{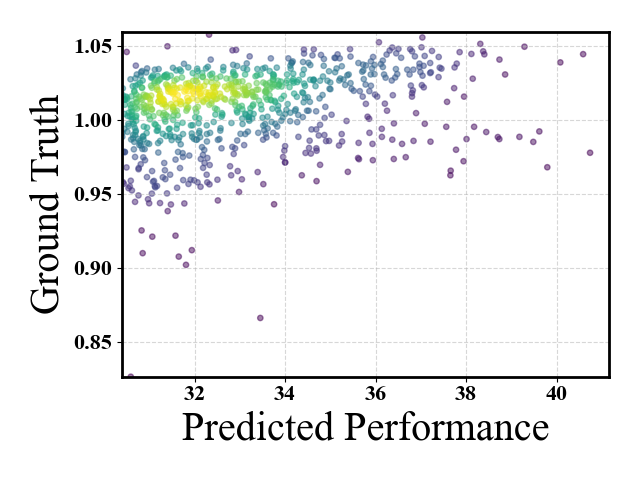}
    \caption{NB201: ImageNet16 (Top 5\%)}
  \end{subfigure}
  
  \caption{Comparison of neural architecture search (NAS) results on different datasets using NAS-Bench-101 and NAS-Bench-201. Figures (a) to (d) show a scatterplot of architecture performance on NAS-Bench-101 and NAS-Bench-201 for CIFAR-10, CIFAR-100, and ImageNet16-120 datasets. Figures (e) to (h) display the top-performing architectures on each dataset using NAS-Bench-101 and NAS-Bench-201.}\label{fig:all_corr_nb101_201}
\end{figure*}

We provide the correlation of ParZC across multiple datasets and benchmarks in Figure~\ref{fig:all_corr_nb101_201}. The figures present a comparison of neural architecture search (NAS) results on different datasets using two benchmark datasets: NAS-Bench-101 and NAS-Bench-201.
Figures (a) to (d) show scatterplots illustrating the performance of various architectures on NAS-Bench-101 and NAS-Bench-201 for CIFAR-10, CIFAR-100, and ImageNet16-120 datasets. These scatterplots provide insights into the distribution and characteristics of the architectures across different datasets. 
Figures (e) to (h) showcase the top-performing architectures found through NAS-Bench-101 and NAS-Bench-201 on each dataset, highlighting the architectures that achieved the highest performance. These figures demonstrate the effectiveness and potential of neural architecture search in discovering architectures optimized for specific datasets.

{
    \small
    \bibliographystyle{ieeenat_fullname}
    \bibliography{main}
}


\end{document}